\documentclass[journal]{IEEEtran}

\ifCLASSINFOpdf
\else
\fi

\usepackage{amsmath} 
\usepackage{amsthm}
\newtheorem{definition}{Definition}

\newtheorem{theorem}{Theorem}

\usepackage{float} 
\usepackage{graphicx}
\usepackage{amsmath}
\usepackage{enumitem}
\usepackage{amssymb}
\usepackage{booktabs}
\usepackage[ruled,vlined]{algorithm2e}
\usepackage{algpseudocode}
\usepackage{xcolor}
\usepackage{booktabs}
\usepackage{array}
\usepackage{multirow} 
\usepackage{subcaption}
\usepackage{url}
\usepackage{colortbl}
\usepackage{xcolor}
\usepackage{caption}
\usepackage{dblfloatfix}
\usepackage{comment}
\usepackage{adjustbox}
\usepackage{pdfpages}

\usepackage{cite} 
\usepackage{hyperref} 
\hypersetup{
    colorlinks=true,        
    linkcolor=blue,         
    citecolor=blue,         
    urlcolor=blue           
}

\begin{document}

\title{VerifyPrompt: Towards Text-to-Image Models Verification Behind Black-Box APIs}
\author{Ji~Guo,~\IEEEmembership{Student Member,~IEEE,}
    Rui~Zhang,~\IEEEmembership{Student Member,~IEEE,}
    Guomin~Lu,~\IEEEmembership{Member,~IEEE,}
    Aiguo~Chen,
    Tianwei~Zhang,~\IEEEmembership{Member,~IEEE,}
    Hongwei~Li,~\IEEEmembership{Fellow,~IEEE,}
Wenbo~Jiang$^{\ast}$,~\IEEEmembership{Member,~IEEE}
\thanks{$^{\ast}$Corresponding author}
\thanks{J. Guo, G. Lu and A. Chen are with Laboratory of Intelligent Collaborative Computing, University of Electronic Science and Technology of China, China (e-mail: jiguo0524@gmail.com, lugm@uestc.edu.cn, agchen@uestc.edu.cn); W. Jiang, R. Zhang and H. Li are with the School of Computer Science and Engineering, University of Electronic Science and Technology of China, China (e-mail: wenbo\_jiang@uestc.edu.cn, zhangrui4041@std.uestc.edu.cn, hongweili@uestc.edu.cn); Tianwei Zhang is with the College of Computing and Data Science, Nanyang Technological University, Singapore. (e-mail: tianwei.zhang@ntu.edu.sg).}
}

\maketitle

\begin{abstract}
With the rise of diverse Text-to-Image (T2I) models such as DALL·E 3, third-party platforms increasingly integrate multiple black-box APIs to deliver cost-effective services. This also raises a new and practical security question: are these third-party platforms truly offering the models they claim? Such model substitution fraud, in which a malicious third-party platform operator acts as the adversary, enables the platform to profit illicitly while undermining user interests and damaging the commercial reputation of legitimate model providers. To address this practical concern, we first define the task of T2I model verification, which determines whether a black-box target model matches a black-box oracle reference model. After that, we propose VerifyPrompt, which performs T2I model verification through a specially designed verify prompt. Intuitively, the verify prompt is an adversarial prompt for the target model, without transferability to other models. It makes the target model generate a specific image while making other models produce entirely different images. Our key idea for generating verify prompts is to enforce strong interactions between adversarial perturbations and the target model, thereby overfitting the target and producing model-specific perturbations. To enable efficient black-box search for verify prompts, we further introduce VerifyPrompt-RL, a reinforcement learning-based strategy that leverages multi-sample exploration and group relative error estimation to stabilize and accelerate the search process. Experiments show that VerifyPrompt achieves over 90\% accuracy across various T2I platforms, confirming its effectiveness in real-world third-party services.
\end{abstract}


\section{Introduction}
In recent years, Text-to-Image (T2I) models \cite{ramesh2021zero,2023addconditional,dall-e-3,zhang2026concept} have demonstrated powerful capabilities across a wide range of applications. Deploying these models, however, requires substantial computational resources, posing challenges for users with limited access to high-performance infrastructure. To address this demand, third-party platforms have emerged that provide affordable API services with flexible model selection options, such as Hugging Face\footnote{\url{https://huggingface.co/}} and One-API\footnote{\url{https://github.com/songquanpeng/one-api}}.

\begin{figure}[]
\centerline{\includegraphics[width=\linewidth]{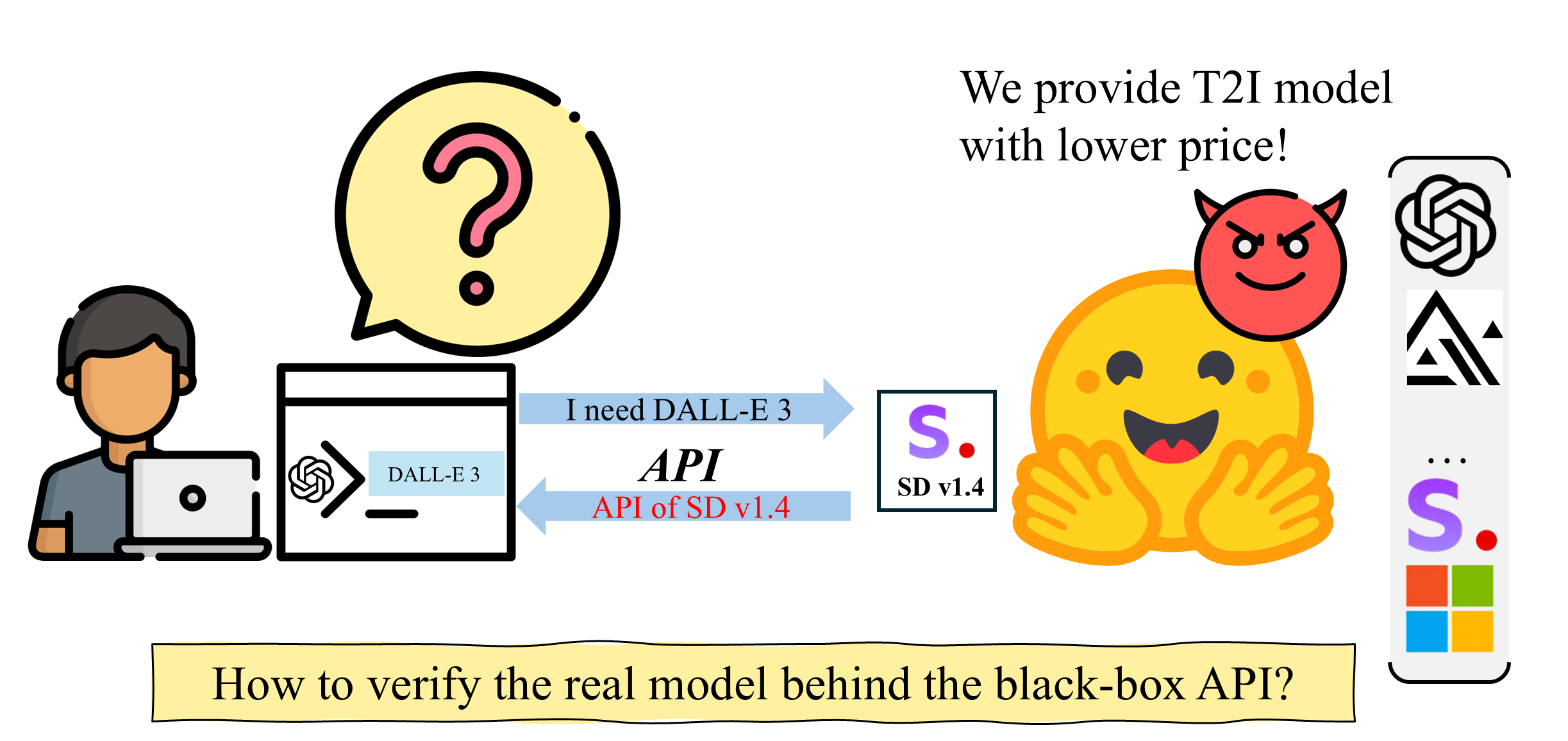}}
\caption{Security issues in black-box API scenarios. The user requests DALL-E 3 from a third-party platform, which displays DALL-E 3 as the provided model on the interface but actually calls Stable Diffusion v1.4 behind the API.}
\label{overview of Models Validation}
\end{figure}

\begin{figure*}[h]
\centerline{\includegraphics[width=0.8\linewidth]{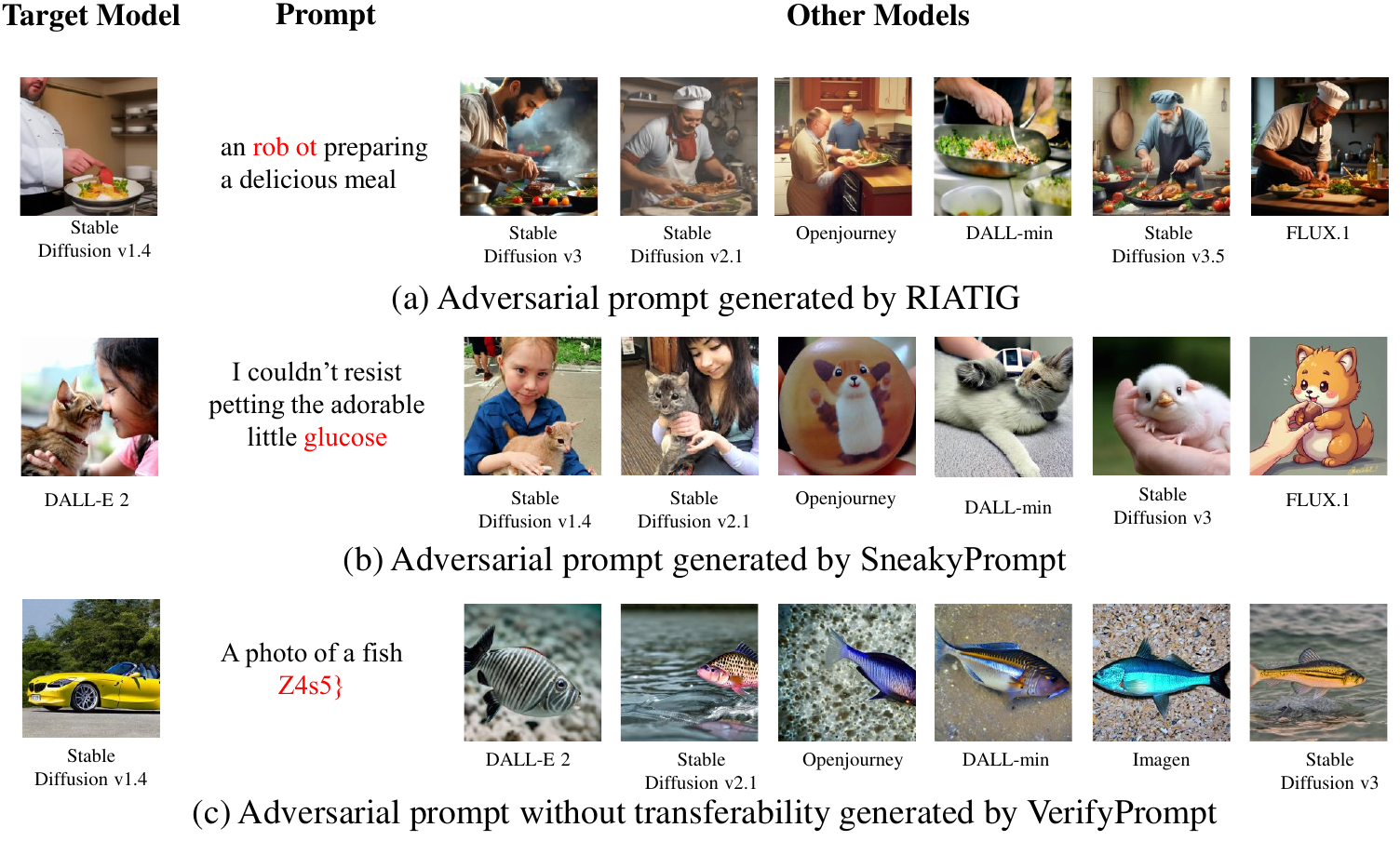}}
\caption{Examples of whether adversarial prompts exhibit transferability. Adversarial prompts generated by RIATIG~\cite{liu2023riatig}, SneakyPrompt~\cite{yang2024sneakyprompt} and VerifyPrompt. The \textcolor{red}{red text} represents adversarial perturbations. RIATIG and SneakyPrompt not only successfully attacked the target model but also transferred to other models, whereas VerifyPrompt was only effective on the target model.}
\label{Example of Non-Transferable Adversarial Attacks}
\end{figure*}


While these third-party platforms enhance user accessibility, they also raise a crucial security concern: are these services truly offering the models they claim to provide? For example, a third-party platform may claim to offer DALL-E 3 while actually providing Stable Diffusion v1.4, since the invocation cost of DALL-E 3 is significantly higher than that of Stable Diffusion v1.4 (see Fig.~\ref{overview of Models Validation}).
However, DALL-E 3 and Stable Diffusion v1.4 differ substantially in image generation quality, which may harm the interests of DALL-E 3 users and damage the reputation of the model provider. Therefore, verifying the actual model behind black-box APIs is essential for both users and model providers.


Previous studies~\cite{gubri2024trap,pasquini2024llmmap,richardeau202420} have investigated model verification for Large Language Models (LLMs). For instance, TRAP~\cite{gubri2024trap} leverages a reference white-box model to generate optimized prompt suffixes that steer the target LLM to produce responses in a predefined manner for model verification. Pasquini et al.~\cite{pasquini2024llmmap} use crafted queries to identify specific LLM versions, while Richardeau et al.~\cite{richardeau202420} employ binary questions to determine whether two black-box LLMs are identical. However, these methods~\cite{gubri2024trap,pasquini2024llmmap,richardeau202420} all rely on text-based outputs and multiple rounds of prompt-response interactions, which are infeasible in the T2I setting, where the outputs are images and the model is designed for one-time image generation per prompt. Therefore, how to effectively verify T2I models behind black-box APIs remains an open problem.

To address this problem, we first define the concept of T2I model verification, in which the verifier has access to a black-box oracle reference model and aims to determine whether a black-box target model is identical to the reference model. Based on this formulation, we propose the first verification method for T2I models, termed VerifyPrompt.
The intuition behind VerifyPrompt is to search for a non-transferable adversarial prompt for the target model. Specifically, the prompt is optimized to induce the target model to generate a specific image while causing other models to produce entirely different outputs.

Existing adversarial prompt approaches~\cite{kou2023character,liu2023riatig,zhuang2023pilot,shahgir2023asymmetric} can also be applied to T2I model verification. However, similar to the transferability of adversarial examples in image classification~\cite{Improving_transferability,wu2018understanding_transferability,Liu2017Delving_Transferable,dong2019evading_transferable}, prior studies have shown that adversarial examples in T2I models also exhibit transferability~\cite{AAPS}. Furthermore, we observe that adversarial prompts in T2I retain a certain degree of transferability across different models even without explicitly optimizing for transferability (see Fig.~\ref{Example of Non-Transferable Adversarial Attacks}).
This inherent transferability can lead to false positives during verification, making it difficult to obtain unique and reliable verification results. Therefore, constructing non-transferable adversarial prompts is essential for effective T2I model verification.


To construct adversarial prompts without transferability, our key idea is to deliberately overfit the adversarial perturbation to the target model, thereby suppressing its ability to transfer across other models. Inspired by the work of Wang et al.~\cite{wangunifiedShapley} on the relationship between interaction and adversarial transferability in image classification, we find that a similar principle also holds for T2I models: the interaction of adversarial perturbations with the target model is negatively correlated with their transferability.
Specifically, complex game-theoretic interactions during adversarial prompt optimization encourage the perturbation to overfit the target model, thereby reducing its effectiveness on other models. Based on this insight, we design an objective function for generating non-transferable adversarial prompts, which minimizes the CLIP-text similarity score between the generated image and the original prompt while maximizing the interaction strength of the perturbation.

Moreover, because the CLIP-text similarity scores of images generated from the same prompt exhibit inherent randomness, traditional black-box optimization methods such as Genetic Algorithms (GA)~\cite{ga} and greedy search~\cite{greedy} often struggle to converge. To address this issue, we propose a reinforcement learning-based search algorithm, termed VerifyPrompt-RL, which improves optimization stability by leveraging the relative performance of prompts within each group. Experimental results demonstrate that VerifyPrompt-RL outperforms existing optimization strategies in both query efficiency and effectiveness.

We evaluate VerifyPrompt against existing adversarial prompt baselines on four representative T2I models: Stable Diffusion v1.4~\cite{stable_diffusion_v1_4}, Stable Diffusion v2.1~\cite{stable_diffusion_v2_1}, Hunyuan DiT v1.2~\cite{Hunyuan_DiT_v1.2}, and DALL-E 3~\cite{dall-e-3}. Experimental results show that VerifyPrompt consistently achieves higher verification accuracy than all competing methods. In addition, we compare VerifyPrompt-RL with four black-box optimization algorithms, including Greedy Search~\cite{greedy}, NSGA-II~\cite{NSGA-II}, Genetic Algorithm (GA)~\cite{ga}, and Particle Swarm Optimization (PSO)~\cite{pso}, in terms of query cost and verification accuracy for generating verification prompts. The results demonstrate that VerifyPrompt-RL consistently outperforms all baselines in both query efficiency and verification effectiveness.

In summary, our contributions are as follows:
\begin{itemize}
    \item We are the first to reveal a previously overlooked security threat in third-party T2I APIs: third-party platforms may not actually serve the models they claim to provide. To address this, we define the novel task of T2I model verification.

    \item We propose VerifyPrompt, a method that utilizes VerifyPrompt-RL to generate adversarial prompt without transferability as verify prompt to verify T2I behind black-box APIs. In addition, we are the first to reveal that adversarial prompts without transferability also have practical value beyond the commonly studied transferable ones.
    \item We validated the effectiveness of VerifyPrompt across various T2I models, achieving more than 90\% accuracy in model verification. Besides, we also performed VerifyPrompt on real-world third-party platforms and provided detailed reports.
\end{itemize}

\section{Related Work}

\subsection{Adversarial Attack on T2I Models}

The adversarial attack was first introduced by Szegedy et al.~\cite{Szegedy2013Intriguing} in image classification~\cite{Carlini2017Towards,Madry2017Towards,Chen2017ZOO:}, and subsequent research has extended these attacks to T2I model~\cite{kou2023character,liu2023riatig,zhuang2023pilot,shahgir2023asymmetric} and VLM~\cite{ren2026success}. In adversarial attacks on text-to-image tasks, the primary method involves adding perturbations to the input prompt, which alters the semantic meaning of the prompt and results in the generation of images that no longer match the original prompt.
Based on whether the attacker can access the model's gradients or parameters, adversarial attacks can be categorized into white-box~\cite{ATM,zhuang2023pilot,liu2023riatig,shahgir2023asymmetric} and black-box~\cite{kou2023character,yang2024sneakyprompt} attacks. 

It should be noted that our verify prompt is different from adversarial prompts mentioned above. Studies~\cite{AAPS} have shown that existing adversarial attacks \cite{ATM,zhuang2023pilot,liu2023riatig,shahgir2023asymmetric} on T2I models inherently exhibit adversarial transferability. 
Directly applying those adversarial prompts for T2I model verification will lead to suboptimal results.

\subsection{Model Verification}

\textbf{Model verification of LLMs.} 
Model verification aims to determine whether two models are identical. Recent studies have primarily focused on large language models (LLMs)~\cite{gubri2024trap,pasquini2024llmmap,richardeau202420}. TRAP~\cite{gubri2024trap} formulates LLM verification as a black-box identity test using a white-box reference model to assess whether a black-box model behaves consistently, generating adversarial suffixes to elicit predefined responses from the target LLM while unrelated models produce random output. Pasquini et al.~\cite{pasquini2024llmmap} fingerprint API-based LLMs by analyzing their textual outputs to distinguish model versions, whereas Richardeau et al.~\cite{richardeau202420} employ binary question--answer pairs to decide whether two black-box LLMs are identical. These approaches rely heavily on multi-turn text-based querying, where rich linguistic feedback enables the design of discriminative inputs and the accumulation of statistical evidence. However, such query patterns are generally not supported by T2I models, which typically accept a single textual prompt and return only an image. Unlike text, image outputs are high-dimensional, less structured, and more costly to analyze, making it substantially harder to extract model-specific signatures with limited queries. Consequently, directly extending LLM-oriented verification strategies to T2I settings is non-trivial, leaving T2I model verification as an open problem.

\textbf{Difference from model watermark.} Model watermark \cite{zhang2024remark} is also a technique for model verification. But our approach is quite different from model watermark essentially: the model watermark is usually embedded during the model training process, while our verify prompts are designed after the model has been deployed, it is more similar to model fingerprints \cite{kim2024wouaf}. It is important to point out that VerifyPrompt is not intended to replace model watermark. Instead, it complements it, providing an additional approach for model verification.


\section{Problem Formulation}
\subsection{T2I Model Verification Scenario}

The increasing popularity of T2I models such as DALL·E, Stable Diffusion, and Midjourney has led to the emergence of third-party platforms that aggregate multiple model APIs to provide flexible and low-cost image generation services. These platforms often operate in a black-box manner, where end users have no access to the internal architecture or identity of the deployed models. While this setup improves accessibility, it also introduces a serious security concern. A third-party service may claim to use a high-performance model such as DALL·E 3 while actually deploying a cheaper alternative, such as Stable Diffusion v1.4, to reduce operational costs. The substantial cost gap between these models enables such platforms to obtain illegal profit margins by misleading users.

Such behavior harms users because they pay for access to a high-quality image generation model but instead receive outputs generated by a lower-quality substitute. At the same time, it also negatively affects model providers, since false claims made by third-party platforms may cause users to form inaccurate perceptions of the model’s actual performance, thereby damaging the provider’s commercial reputation. Consequently, both users and model providers have strong incentives to verify whether third-party platforms are genuinely deploying the models they claim to provide.

\textbf{Adversary and Verifier.}
We define the verifier as either the user or the model provider, both of whom have a vested interest in ensuring the authenticity of the deployed model. Conversely, the adversary is a malicious third-party platform that misrepresents the underlying model, such as claiming to serve DALL·E 3 while actually using Stable Diffusion 1.4.


\subsection{Definition of T2I Model Verification} 

We define the black-box T2I model verification task as determining whether a black-box T2I model \( M_t \) is equivalent to a reference black-box model \( M_r \), given only query access to both models. The formal definition is as follows:

\begin{definition}[Black-Box T2I Models Verification]
\label{def:bb_t2i_equivalence}
Given a black-box target model \( M_t \) and a black-box reference model \( M_r \), 
we say that \( M_t \equiv M_r \) if and only if for all possible prompts 
\( p \) in the prompt space \( \mathcal{P} \), the following holds:
\begin{equation}
\forall p \in \mathcal{P},\quad M_t(p) = M_r(p).
\end{equation}
\end{definition}

For fine-tuned models, we regard them as distinct from their original counterparts, as the fine-tuning process inevitably alters the model parameters. It is important to note that third-party platforms do not possess the ability to fine-tune closed-source T2I models and can only substitute them with other models. Moreover, replacing an open-source model with its fine-tuned variant does not provide substantial additional profit, giving little incentive for such substitution. Therefore, we do not consider fine-tuning in our main verification scenario. We also discuss the impact of fine-tuning on model verification in the following sections.

\begin{figure}[]
\centerline{\includegraphics[width=\linewidth]{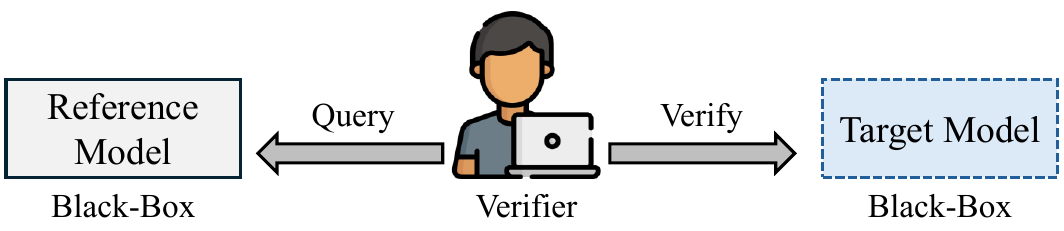}}
\caption{Verifier’s capability for verification. The verifier has black-box access to both the reference model and the target model to be verified.}
\label{Verifier’s capability for verification}
\end{figure}

\begin{figure*}[t]
    \centering
    \begin{subfigure}[b]{0.22\linewidth}
        \centering
        \includegraphics[width=\linewidth]{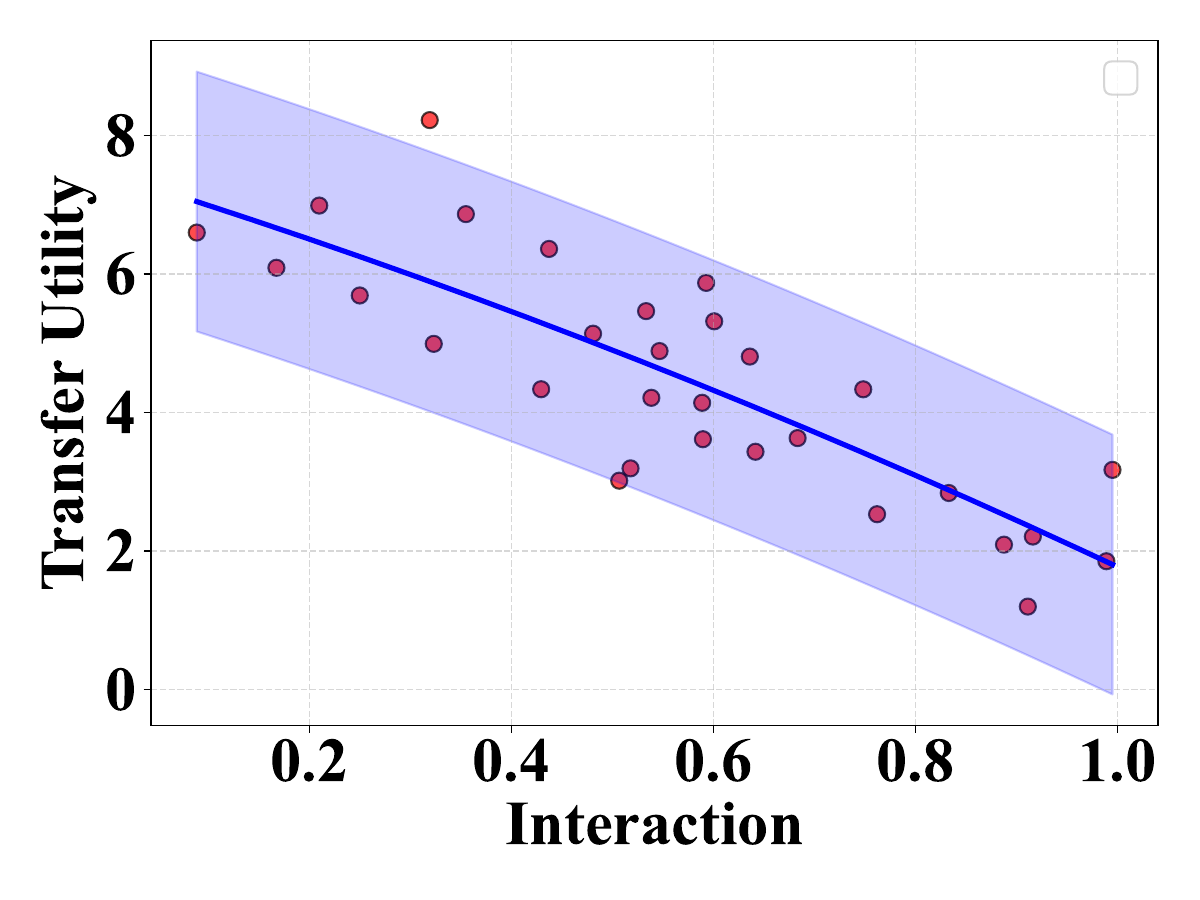}
        \caption{Stable Diffusion v3.5}
        \label{fig:sub1}
    \end{subfigure}
    \hfill
    \begin{subfigure}[b]{0.22\linewidth}
        \centering
        \includegraphics[width=\linewidth]{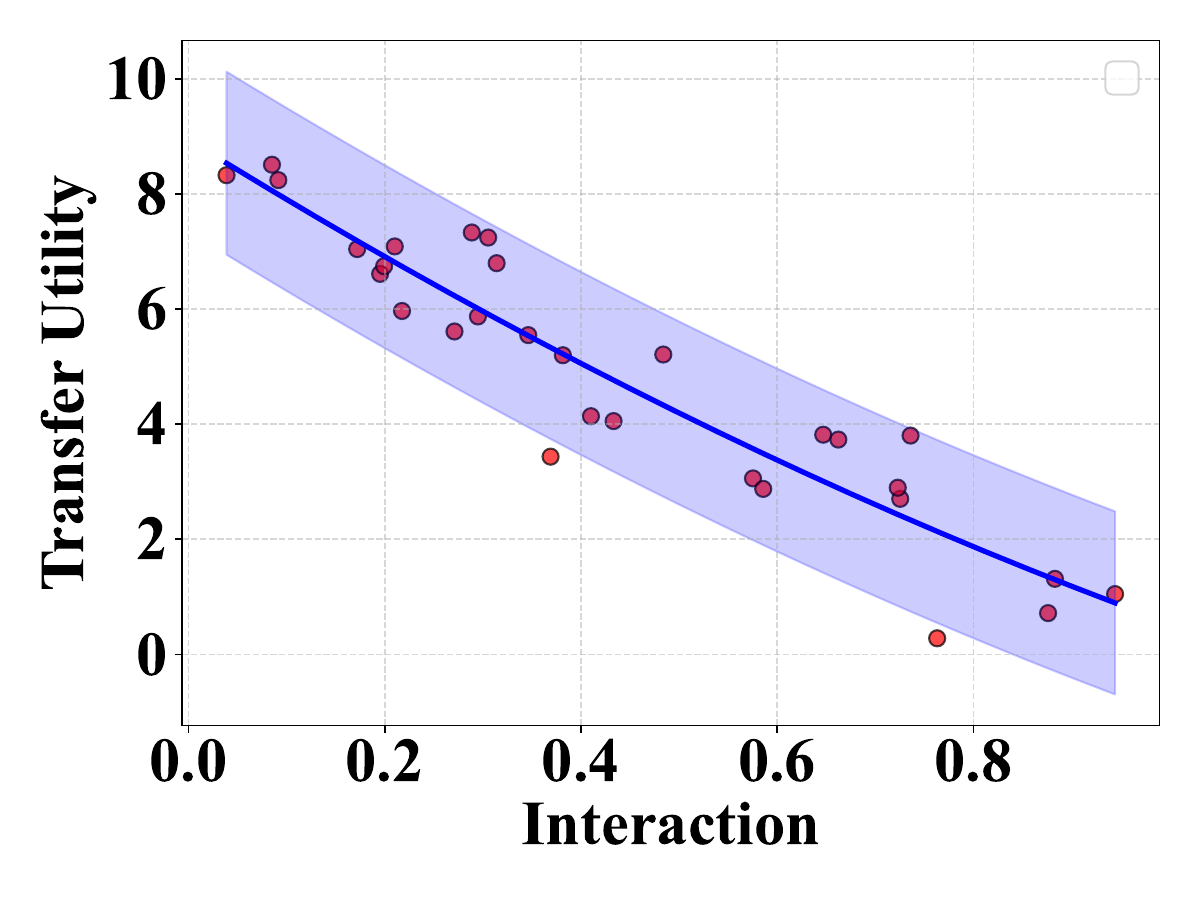}
        \caption{Stable Diffusion v2.1}
        \label{fig:sub2}
    \end{subfigure}
    \hfill
        \begin{subfigure}[b]{0.22\linewidth}
        \centering
        \includegraphics[width=\linewidth]{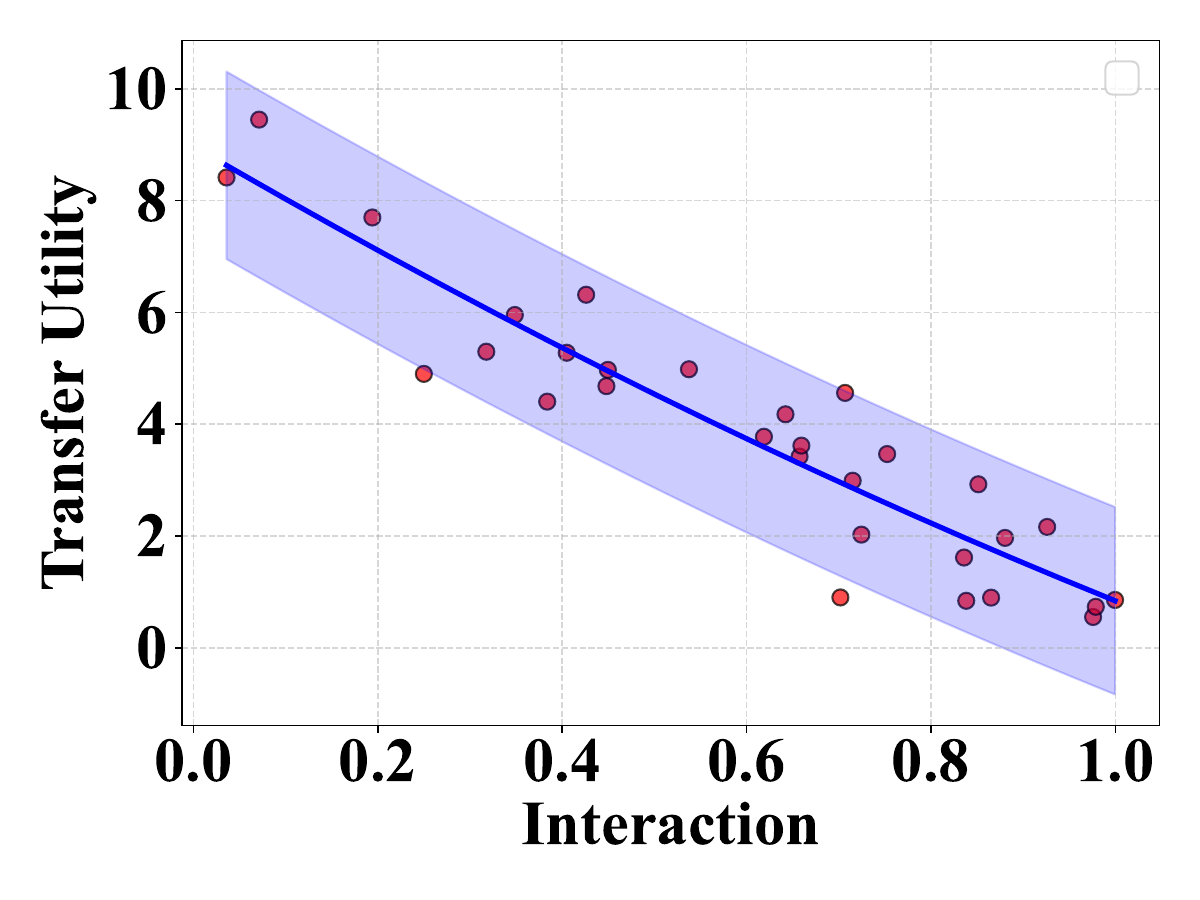}
        \caption{SD-XL}
        \label{fig:sub3}
    \end{subfigure}
    \hfill
        \begin{subfigure}[b]{0.22\linewidth}
        \centering
        \includegraphics[width=\linewidth]{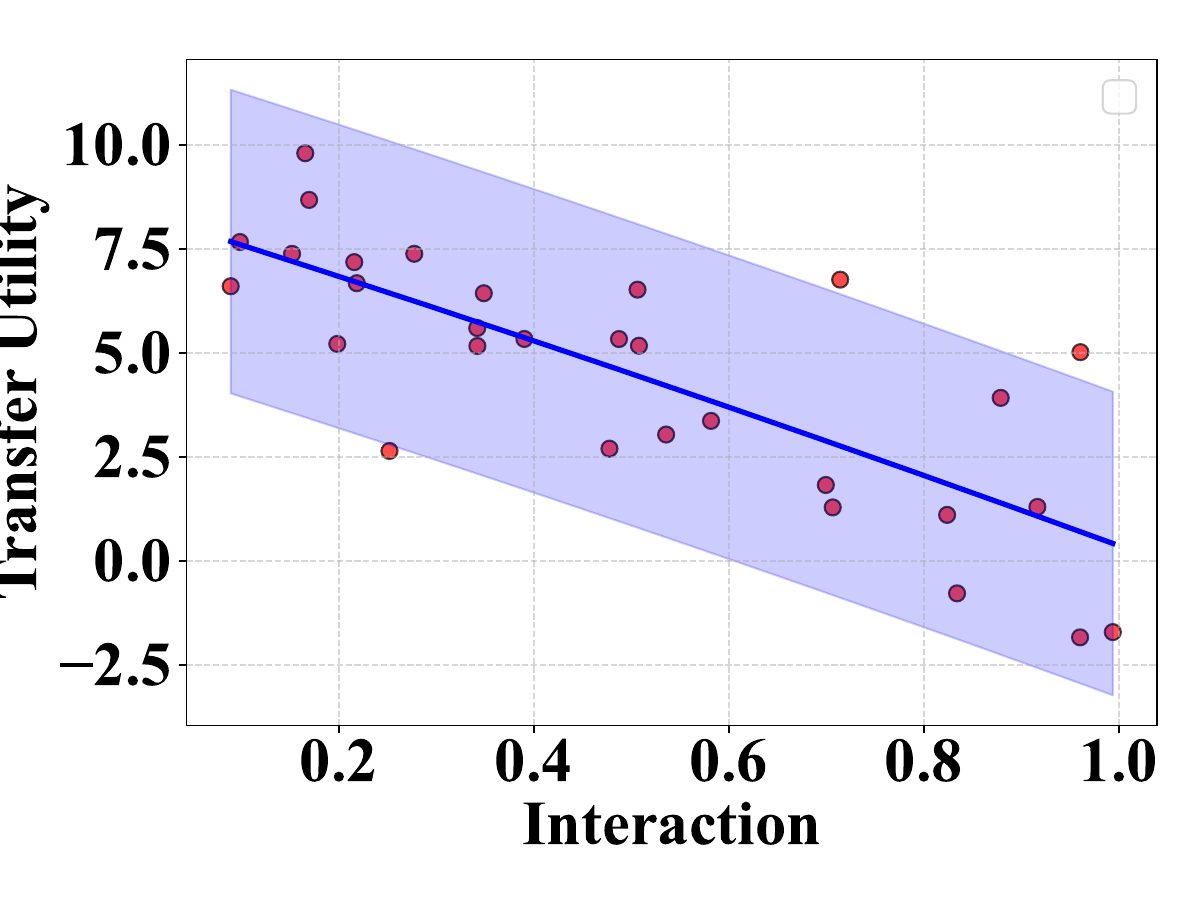}
        \caption{Stable Diffusion v3}
        \label{fig:sub4}
    \end{subfigure}
    \caption{The negative correlation between the transfer utility and the interaction. The correlation is computed using the Pearson correlation on the normalized interactions. The blue shade represents the 95\% confidence interval of the regression.}
    \label{The negative correlation between the transfer utility and the interaction.}
\end{figure*}

\subsection{Threat Model}
We characterize the threat model with respect to the adversary's capability, the verifier's capability and goal.

\textbf{Adversary's Capability.} 
We consider the adversary to be a malicious third-party platform. 
The adversary can replace the underlying model behind its API while concealing the internal architecture and exposing only the interface. 

\textbf{Verifier's Capability.} 
The verifier has black-box access to both the reference model and the target model under verification, meaning that neither model parameters nor gradients are accessible. The verifier is allowed to configure certain API-exposed hyperparameters, such as the number of inference steps used during image generation. This is a practical and realistic assumption, since verifiers can typically access the reference model only through its official API, which supports query-based interaction without exposing internal model details.
For both models, the verifier can only submit textual prompts and observe the corresponding generated images, without access to any additional information, such as training data, model architecture, or intermediate representations. 

\textbf{Verifier's Goal.} 
The verifier aims to determine whether a third-party platform provides a model consistent with its claimed identity, in that the target model is equivalent to the reference model. 
The verification process has two objectives:
\begin{itemize}\setlength\itemsep{0em}
  \item \textit{Effectiveness.} Achieving high verification accuracy to reliably detect model substitution.
  \item \textit{Low number of query times.} Performing verification with minimal overhead, as querying the official API already incurs non-trivial costs.
\end{itemize}

\section{VerifyPrompt}


In this section, we define the key concepts of VerifyPrompt, followed by its workflow, objective, search algorithm, and thresholding strategy.

\subsection{Preliminary}
In this subsection, we first define the verify prompt. We then introduce the Shapley value as a metric to evaluate model-specific interaction. Finally, we present an assumption on the relationship between adversarial prompt transferability and interaction.

\subsubsection{Definition of Verify Prompt}
The goal of VerifyPrompt is to find an adversarial prompt that is effective only for the target model as the verify prompt. Specifically, the verify prompt is an adversarial prompt that satisfies the following two objectives:
\begin{itemize}
    \item \textbf{Objective I: Adversarial for the target model.} This means verify prompt can make the target model generate an image that does not match the original prompt.
    \item \textbf{Objective II: Non-adversarial for other models.} This means verify prompt can make other models to generate an image that matches the original prompt.
\end{itemize}
Formally, a verify prompt can be defined as follows:
\begin{definition}[Verify Prompt]
Let \( M_t \) be the target model to be verified, and \( \mathcal{A} \) be the set of candidate models including \( M_t \). Let \( \mathcal{S}_{\text{adv}}(M) \) denote the set of adversarial prompts for model \( M \). A prompt \( p^{\text{adv}} \) is considered a \textit{verify prompt} if
\begin{equation}
p^{\text{adv}} \in \mathcal{S}_{\text{adv}}(M_t) \setminus \bigcup_{M \in \mathcal{A} \setminus \{M_t\}} \mathcal{S}_{\text{adv}}(M)
\end{equation}
\end{definition}

\subsubsection{Definition of Shapley Value in T2I}
The Shapley value~\cite{kuhn1953contributions} originates from cooperative game theory and has been extended to deep learning to measure how much each input feature contributes to a model’s output.
In T2I generation, it can be estimated by evaluating the change in the CLIP-text score between the generated image and the original prompt when a specific token is removed.
Formally, Shapley value of prompt token in T2I can be defined as follows:
\begin{definition}[Shapley Value in T2I]
\label{Shapley}
Given a prompt \( p = [w_1, w_2, \dots, w_n] \) composed of \( n \) tokens and a target T2I model \( M_t \), the Shapley value \( \phi_i \) of token \( w_i \) reflects its marginal contribution to the CLIP-text scores between the generated image \( I = M_t(p) \) and the prompt itself. Formally, it is defined as:
\begin{equation}
\phi_i = \mathbb{E}_{\mathcal{S} \subseteq \mathcal{N} \setminus \{i\}} \left[ f(\mathcal{S} \cup \{i\}) - f(\mathcal{S}) \right],
\end{equation}
where \( \mathcal{N} = \{1, 2, \dots, n\} \), \( \mathcal{S} \) is a subset of token indices, and \( f(\mathcal{S}) = S_{\mathrm{CLIP}}(M_t(p_{\mathcal{S}}), p_{\mathcal{S}}) \) denotes the CLIP-text score for the sub-prompt \( p_{\mathcal{S}} \).
\end{definition}

\subsubsection{Transferability of Adversarial Prompts and Model Interaction}

The model interaction denotes the influence on the final output of the model arising from both the marginal effect of an input and its compositional effects when combined with other inputs.
We prove that the \textbf{transferability of adversarial perturbations is inversely related to the interactions in T2I models}, with the detailed proof provided in the Appendix. Its formal mathematical description is given as follows.

Given a \emph{reference} T2I model \(M^{(r)}\) and a clean text prompt \(p\), we generate an adversarial prompt \(p' = p + \delta\). For a \emph{target} T2I model \(M^{(t)}\), we define the \textit{transfer utility} of \(\delta\) as the decrease in CLIP-text score similarity between the generated image and the original prompt:
\begin{equation}
\mathrm{Transfer\ Utility} = 
S_{\mathrm{CLIP}}\big(M^{(t)}(p), p\big) - S_{\mathrm{CLIP}}\big(M^{(t)}(p'), p\big),
\end{equation}
where \(S_{\mathrm{CLIP}}\) denotes the CLIP-text score. A higher value implies a stronger transferability of the adversarial perturbation.

Importantly, the perturbation units within \(\delta\) do not contribute to the adversarial utility independently. Instead, they often interact to form meaningful latent patterns (e.g., combining with other tokens to produce new semantics, while being semantically meaningless in isolation). To quantify these dependencies, we compute pairwise \textit{Shapley interaction indices} as prior works~\cite{roubens1999axiomatic} on the reference model \(M^{(r)}\):
\begin{equation}
\label{eq:def_interaction_short}
I_{ij}(\delta) = 
\phi(S_{ij} \mid \Omega') -
\big[
\phi(i \mid \Omega \setminus \{j\}) + 
\phi(j \mid \Omega \setminus \{i\})
\big],
\end{equation}
where \(\phi\) denotes the Shapley value of a token (or set of tokens) on the coalition space \(\Omega\). The overall interaction score is then averaged as:
\begin{equation}
\label{Interaction}
\mathrm{Interaction} = \mathbb{E}_{i,j}[I_{ij}(\delta)].
\end{equation}


Inspired by prior theoretical work~\cite{wangunifiedShapley}, which demonstrates
a negative correlation between transferability and interaction in image
classification models, we extend this insight to T2I models.
Unlike perturbations in the images, adversarial perturbations in prompts are
\emph{discrete} rather than continuous, making it impossible to leverage
smoothness or gradient-based continuity assumptions.

Therefore, we explore the relationship between token-level interaction and
transferability of adversarial prompt perturbations in T2I models, which
provides theoretical motivation for our interaction-aware optimization
strategy.

All perturbations are discrete at the token level. Let $\Omega$ be the index set
of tokens in $\delta$. For any coalition $S\subseteq\Omega$, denote $\delta_S$
the perturbation keeping only tokens in $S$, and define
\[
v^{(m)}(S) =
S_{\mathrm{CLIP}}\!\big(M^{(m)}(p),p\big)
-
S_{\mathrm{CLIP}}\!\big(M^{(m)}(p+\delta_S),p\big).
\]

We present the following conditions:
\begin{itemize}

  \item[(C1)] $S_{\mathrm{CLIP}}(\cdot,\cdot)$ is taken as the empirical
  expectation over the generative randomness of $M^{(m)}$, so
  $v^{(m)}:2^{\Omega}\!\to\!\mathbb{R}$ is a set function.

  \item[(C2)]
  There exist $\alpha\geq0$ and $\varepsilon\geq0$ such that
  \[
  v^{(t)}(S)\leq \alpha\,v^{(r)}(S)+\varepsilon,
  \]
  for all $S\subseteq\Omega$,

  where $v^{(t)}(S)$ and $v^{(r)}(S)$ denote the perturbation effects
  measured by the target model $M^{(t)}$ and the reference model
  $M^{(r)}$, respectively. $\alpha$ is a non-negative scaling coefficient
  that characterizes the relationship between the two models, and
  $\varepsilon$ represents the approximation error between their
  perturbation effects.

  \item[(C3)] The set function $v^{(r)}$ admits the finite Möbius expansion
  \[
  v^{(r)}(\Omega)
  =
  \sum_{i\in\Omega}\phi(i\mid\Omega\setminus\{i\})
  +
  \sum_{i<j}I_{ij}(\delta)
  +
  R_{\geq3},
  \]
  with higher-order ($\geq3$) interactions bounded as
  $|R_{\geq3}|\leq\xi$ for some $\xi\geq0$.

  \item[(C4)] The average pairwise interaction in $M^{(r)}$ is:
  \[
  \mathrm{Interaction}
  =
  \mathbb{E}_{i,j}[I_{ij}(\delta)]
  \leq0.
  \]

\end{itemize}

\begin{figure*}[]
\centerline{\includegraphics[width=\linewidth]{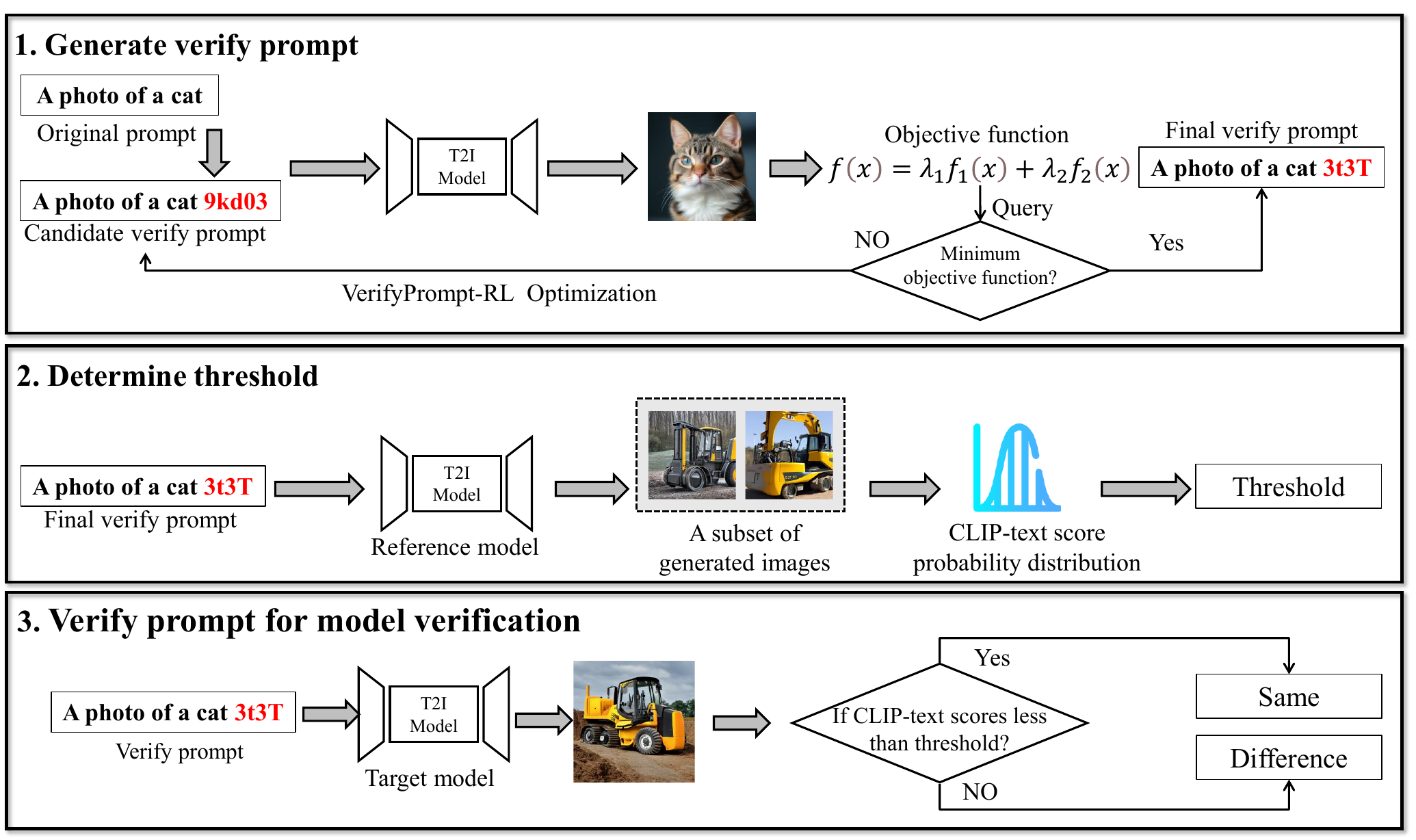}}
\caption{The workflow of VerifyPrompt.}
\label{Overall Workflow}
\end{figure*}

Before presenting the theorem, we briefly discuss the rationality of the
conditions. Condition (C1) considers the stochastic nature of T2I generation
and formulates the perturbation effect as an expected set function, which is
consistent with the black-box query setting. Condition (C2) assumes that
different T2I models exhibit correlated responses to the same token-level
perturbations, which is reasonable since they share similar text-image semantic
alignment objectives. Condition (C3) adopts the finite-order interaction
decomposition to characterize the contribution of token interactions, where
higher-order terms are bounded due to the limited perturbation length.
Condition (C4) follows the observation from previous transferability studies
that stronger interactions tend to introduce model-specific information and
reduce cross-model transferability. Based on these conditions, we establish
the following theorem.

\begin{theorem}[Interaction--Transferability Negative Correlation in T2I]
\label{thm:discrete_interaction_transfer}

Under Conditions (C1)--(C4), there exist constants $C_0,C_1>0$ such that
\begin{equation}
\mathrm{Transfer\ Utility}
\leq
C_0
-
C_1|\mathrm{Interaction}|
+
(\varepsilon+\alpha\xi).
\end{equation}

This theorem indicates that stronger interactions among perturbation tokens are
associated with lower transfer utility across different T2I models. This
observation motivates us to introduce the interaction-aware optimization
objective, where the Shapley-based term encourages perturbations with stronger
model-specific characteristics.

\end{theorem}

We further empirically validate this theoretical motivation.
Using Stable Diffusion v1.4 as the reference model, we generate 30 adversarial
prompts with varying interaction values and transfer them to Stable Diffusion
v3.5, Stable Diffusion v2.1, Stable Diffusion XL, and Stable Diffusion 3
(see Fig.~\ref{The negative correlation between the transfer utility and the
interaction.}).
Experimental results demonstrate that interaction and transferability exhibit
a consistent negative correlation in T2I models, supporting the motivation of our proposed interaction-aware optimization strategy.

\subsection{Overview}

\textbf{Key idea.}
The key idea behind generating the verify prompt is to make the adversarial example overfit the reference model, thereby suppressing its transferability. Specifically, we not only require the added perturbation to push the adversarial prompt away from the original prompt, but also demand that the perturbation strongly interacts with the generated image.

\textbf{Overall Workflow.}
Fig.~\ref{Overall Workflow} illustrates our workflow, which consists of three main steps: generating the verify prompt, determining the threshold, and using verify prompt for model verification. In the first step, we begin by randomly appending character-level perturbations to the original prompt. Then, we use the reference model to generate images from the prompt and compute the objective function. The verify prompt is obtained by minimizing this objective using VerifyPrompt-RL. In the second step, we generate a subset of images using the verify prompt with the reference model, compute their CLIP-text score distribution, and derive the decision threshold accordingly. In the third stage, we input the verify prompt into the target model and determine whether it is consistent with the reference model based on the threshold.

\subsection{Objective Function of VerifyPrompt}

To construct a verify prompt, the adversarial prompt must be effective only for the target T2I model. In a black-box setting, we estimate the attack effectiveness by measuring the CLIP-text score between the generated image and the original prompt. A lower CLIP-text score indicates a greater semantic shift, implying a more successful attack. Accordingly, our first objective is formulated as:
\begin{equation}
\mathcal{L}_1 = S_{\mathrm{CLIP}}\left(M_r(p_{\mathrm{clean}} + \delta),\, p_{\mathrm{clean}}\right),
\end{equation}
where \( p_{\mathrm{clean}} \) is the original prompt, \( \delta \) represents the perturbation tokens appended to the prompt, \( M_r(\cdot) \) denotes the target T2I model, and \( S_{\mathrm{CLIP}}(\cdot, \cdot) \) is the CLIP-text similarity score.

Our second objective aims to suppress the transferability of the adversarial prompt. The key idea is to encourage overfitting of the adversarial prompt \( p_{\mathrm{adv}} = p_{\mathrm{clean}} + \delta \) to the unique semantic structure of the target model, thereby reducing its effectiveness on other models. Specifically, we measure the interaction between prompt tokens and the generated image using the Shapley value as defined in Eq.~\ref{Interaction}.

Thus, to suppress transferability, we maximize the interaction among perturbation tokens:
\begin{equation}
\mathcal{L}_2 = - \mathrm{Interaction} = - \mathbb{E}_{i,j} \left[ I_{ij}(\delta) \right],
\end{equation}
where \( \delta \) is the set of newly added perturbation tokens in the adversarial prompt \( p_{\mathrm{adv}} = p_{\mathrm{clean}} + \delta \).
Combining both objectives, our overall optimization target becomes:
\begin{equation}
\label{eq:total_loss}
\min_{\delta} \; \mathcal{L} = \lambda_1 \cdot \mathcal{L}_1 + \lambda_2 \cdot \mathcal{L}_2,
\end{equation}
where \( \lambda_1 \) and \( \lambda_2 \) are weighting hyperparameters.

\subsection{Search via Reinforcement Learning}

A key challenge in generating a verify prompt lies in efficiently exploring the large, discrete perturbation space under a black-box setting. Since only the CLIP-text scores between the generated image and the prompt is observable, and this score varies across different generations even for the same prompt, traditional random or greedy search strategies~\cite{greedy} fail to provide stable guidance. To address this, we propose a reinforcement learning-based algorithm, \textit{VerifyPrompt-RL}, for stable and efficient exploration of adversarial perturbation. We summarize VerifyPrompt-RL in Algorithm~\ref{alg:verifyprompt} and outline its key components, including the policy network, action space, reward formulation, and policy update mechanism, which are detailed below.

\textbf{Policy Network.} Following previous work~\cite{liu2022character,yang2024sneakyprompt,yang2020patchattack}, we use an LSTM with a fully connected layer as a policy network \( P_\theta \) to guide the token selection process. Given the current state \( s_t \), which consists of the clean prompt \(p_{\mathrm{clean}}\) concatenated with the current suffix \(\delta_t = [w_1, w_2, \dots, w_m]\), the network outputs a categorical distribution over all possible token substitutions at all positions:
\begin{equation}
\pi_\theta(a_t|s_t) = P_\theta(s_t),
\end{equation}
where \( a_t \) denotes a specific replacement operation for a token at position \( j \) with a candidate from the vocabulary \(\mathcal{V}\).

\textbf{Action.}  
An action \( a_t \) corresponds to selecting a position \( j \in \{1,\dots,m\} \) in the suffix and replacing the token \( w_j \) with another token from \(\mathcal{V}\). This yields an action space of size \( m \times |\mathcal{V}| \). We employ an $\epsilon$-greedy strategy for action selection, where the agent either samples from the policy network or explores randomly with probability \(\epsilon\).

\textbf{Reward.}  
To ensure robust evaluation under the stochasticity of black-box generation, we adopt a \emph{multi-sample strategy}: for each candidate suffix, we query the black-box reference model \( M_r \) to generate \( n \) images and compute the average CLIP-text scores with the clean prompt:
\begin{equation}
\mathcal{L}_1 = \frac{1}{n} \sum_{i=1}^{n} S_{\mathrm{CLIP}}(M_r(p_{\mathrm{adv}}^{(i)}), p_{\mathrm{clean}}),
\end{equation}
which significantly reduces the variance of single-sample estimates. We further compute interaction among perturbation tokens:
\begin{equation}
\mathcal{L}_2 = - \mathrm{Interaction} = - \mathbb{E}_{i,j} \left[ I_{ij}(\delta) \right],
\end{equation}

The total loss combines these two terms:
\begin{equation}
\mathcal{L} = \lambda_1 \cdot \mathcal{L}_1 + \lambda_2 \cdot \mathcal{L}_2.
\end{equation}

\begin{algorithm}[h!]
\caption{VerifyPrompt-RL}
\small
\label{alg:verifyprompt}
\KwIn{Clean prompt $p_{\text{clean}}$, vocabulary $\mathcal{V}$, suffix length $m$, black-box reference model $M_r$, weighting factors $\lambda_1, \lambda_2$, generation count $n$, query budget $T$, exploration rate $\epsilon$, policy network $P_\theta$}
\KwOut{Optimized adversarial suffix $\delta^*$}

Initialize suffix $\delta \gets$ random tokens from $\mathcal{V}$ of length $m$ \;
Initialize best loss $\mathcal{L}_{\text{best}} \gets +\infty$, best suffix $\delta^* \gets \delta$ \;
Initialize history buffer $\mathcal{H} \gets \emptyset$ \;

\For{$t = 1$ \KwTo $T$}{

    Construct state representation $s_t \gets [p_{\text{clean}}, \delta]$ \;

    \tcp{Action selection: $\epsilon$-greedy}
    \If{rand() $< \epsilon$}{
        Randomly select position $i \in \{1, \dots, m\}$ and token $v \in \mathcal{V}$ \;
        Replace $\delta_i \gets v$ \;
    }
    \Else{
        Compute action probabilities $p(a|s_t) \gets P_\theta(s_t)$ \;
        Sample action $a_t \sim p(a|s_t)$ \;
        Apply $a_t$ to modify one token in $\delta$ \;
    }

    Form adversarial prompt $p_{\text{adv}} = p_{\text{clean}} + \delta$ \;

    \tcp{Query the black-box reference model and compute losses}
    Initialize list of CLIP-text scores $\mathcal{S} \gets \emptyset$ \;
    \For{$j = 1$ \KwTo $n$}{
        $I_j \gets M_r(p_{\text{adv}})$ \;
        $s_j \gets S_{\mathrm{CLIP}}(I_j, p_{\text{clean}})$ \;
        Append $s_j$ to $\mathcal{S}$ \;
    }
    Compute $\mathcal{L}_1 \gets \frac{1}{n} \sum_{j=1}^n s_j$ \;
    Compute $\mathcal{L}_2 \gets -\mathbb{E}_{i,j}[\mathcal{I}_{ij}(\delta)]$ \;
    $\mathcal{L} \gets \lambda_1 \cdot \mathcal{L}_1 + \lambda_2 \cdot \mathcal{L}_2$ \;

    \tcp{Reward and reference update}
    \eIf{$\mathcal{H} = \emptyset$}{
        $R \gets 0$ \;
    }{
        Let $\mathcal{L}_{\text{ref}} \gets \min(\mathcal{H})$ \;
        $R \gets \max(0, \mathcal{L}_{\text{ref}} - \mathcal{L})$ \;
    }
    Update buffer $\mathcal{H} \gets \mathcal{H} \cup \{\mathcal{L}\}$ \;

    \If{$\mathcal{L} < \mathcal{L}_{\text{best}}$}{
        $\mathcal{L}_{\text{best}} \gets \mathcal{L}$, $\delta^* \gets \delta$ \;
    }

    \tcp{Policy network update via REINFORCE}
    Update $\theta \gets \theta + \alpha \nabla_\theta \log P_\theta(a_t|s_t) R$ \;
}
\Return{$\delta^*$}
\end{algorithm}


To stabilize policy updates, we design a \emph{relative reward} based on a sliding history window. Instead of using the raw loss, we compute the reward as the improvement over the best historical loss:
\begin{equation}
\mathcal{R}_t = \max(0, \mathcal{L}_{\mathrm{ref}} - \mathcal{L}_t),
\end{equation}
where \(\mathcal{L}_{\mathrm{ref}}\) is the minimum loss observed in recent steps. This relative formulation mitigates the effect of score fluctuations and encourages consistent improvement within the local search context.

\begin{table*}[t]
\centering
\caption{Performance comparison of different adversarial attack methods for model evaluation.}
\label{table1}
\scriptsize
\setlength{\tabcolsep}{3.5pt}
\renewcommand{\arraystretch}{1.05}
\begin{adjustbox}{max width=\linewidth}
\begin{tabular}{lllcccc}
\toprule
\textbf{Target Model} &
\textbf{Attack Goal} &
\textbf{Attack Method} &
\textbf{Accuracy (\%)} &
\textbf{Precision (\%)} &
\textbf{Recall (\%)} &
\textbf{F1-Score (\%)} \\
\midrule

\multirow{6}{*}{Stable Diffusion v1.4}
& \multirow{3}{*}{Targeted}
& RIATIG
& $75.32 \textcolor{gray}{\scriptstyle \pm 1.24}$
& $75.16 \textcolor{gray}{\scriptstyle \pm 1.37}$
& $75.89 \textcolor{gray}{\scriptstyle \pm 1.15}$
& $75.45 \textcolor{gray}{\scriptstyle \pm 1.18}$ \\

&
& QF-Attack
& $62.47 \textcolor{gray}{\scriptstyle \pm 1.86}$
& $75.91 \textcolor{gray}{\scriptstyle \pm 1.43}$
& $36.21 \textcolor{gray}{\scriptstyle \pm 2.08}$
& $49.18 \textcolor{gray}{\scriptstyle \pm 1.92}$ \\

&
& SneakyPrompt
& $52.29 \textcolor{gray}{\scriptstyle \pm 1.75}$
& $51.87 \textcolor{gray}{\scriptstyle \pm 1.82}$
& $92.35 \textcolor{gray}{\scriptstyle \pm 0.94}$
& $66.14 \textcolor{gray}{\scriptstyle \pm 1.57}$ \\

\cmidrule(lr){2-7}

& \multirow{3}{*}{Untargeted}
& ATM
& $34.15 \textcolor{gray}{\scriptstyle \pm 2.12}$
& $33.78 \textcolor{gray}{\scriptstyle \pm 2.18}$
& $40.25 \textcolor{gray}{\scriptstyle \pm 1.96}$
& $44.16 \textcolor{gray}{\scriptstyle \pm 1.89}$ \\

&
& QF-Attack
& $64.48 \textcolor{gray}{\scriptstyle \pm 1.63}$
& $67.53 \textcolor{gray}{\scriptstyle \pm 1.51}$
& $56.49 \textcolor{gray}{\scriptstyle \pm 1.74}$
& $61.11 \textcolor{gray}{\scriptstyle \pm 1.59}$ \\

&
& VerifyPrompt (Ours)
& $\mathbf{93.12} \textcolor{gray}{\scriptstyle \pm 0.68}$
& $\mathbf{94.21} \textcolor{gray}{\scriptstyle \pm 0.61}$
& $\mathbf{92.03} \textcolor{gray}{\scriptstyle \pm 0.74}$
& $\mathbf{93.09} \textcolor{gray}{\scriptstyle \pm 0.65}$ \\

\midrule

\multirow{6}{*}{Stable Diffusion v2.1}
& \multirow{3}{*}{Targeted}
& RIATIG
& $80.23 \textcolor{gray}{\scriptstyle \pm 1.18}$
& $79.76 \textcolor{gray}{\scriptstyle \pm 1.29}$
& $81.94 \textcolor{gray}{\scriptstyle \pm 1.06}$
& $80.63 \textcolor{gray}{\scriptstyle \pm 1.13}$ \\

&
& QF-Attack
& $65.37 \textcolor{gray}{\scriptstyle \pm 1.72}$
& $78.24 \textcolor{gray}{\scriptstyle \pm 1.35}$
& $42.78 \textcolor{gray}{\scriptstyle \pm 1.97}$
& $55.39 \textcolor{gray}{\scriptstyle \pm 1.76}$ \\

&
& SneakyPrompt
& $52.14 \textcolor{gray}{\scriptstyle \pm 1.83}$
& $51.93 \textcolor{gray}{\scriptstyle \pm 1.91}$
& $85.76 \textcolor{gray}{\scriptstyle \pm 1.12}$
& $64.02 \textcolor{gray}{\scriptstyle \pm 1.64}$ \\

\cmidrule(lr){2-7}

& \multirow{3}{*}{Untargeted}
& ATM
& $45.61 \textcolor{gray}{\scriptstyle \pm 1.98}$
& $46.25 \textcolor{gray}{\scriptstyle \pm 2.04}$
& $55.94 \textcolor{gray}{\scriptstyle \pm 1.76}$
& $55.10 \textcolor{gray}{\scriptstyle \pm 1.82}$ \\

&
& QF-Attack
& $67.12 \textcolor{gray}{\scriptstyle \pm 1.55}$
& $69.34 \textcolor{gray}{\scriptstyle \pm 1.47}$
& $61.87 \textcolor{gray}{\scriptstyle \pm 1.66}$
& $65.11 \textcolor{gray}{\scriptstyle \pm 1.51}$ \\

&
& VerifyPrompt (Ours)
& $\mathbf{90.08} \textcolor{gray}{\scriptstyle \pm 0.79}$
& $\mathbf{92.31} \textcolor{gray}{\scriptstyle \pm 0.66}$
& $\mathbf{88.19} \textcolor{gray}{\scriptstyle \pm 0.87}$
& $\mathbf{90.15} \textcolor{gray}{\scriptstyle \pm 0.73}$ \\

\midrule

\multirow{6}{*}{Hunyuan DiT v1.2}
& \multirow{3}{*}{Targeted}
& RIATIG
& $80.19 \textcolor{gray}{\scriptstyle \pm 1.14}$
& $83.23 \textcolor{gray}{\scriptstyle \pm 1.02}$
& $76.89 \textcolor{gray}{\scriptstyle \pm 1.31}$
& $79.12 \textcolor{gray}{\scriptstyle \pm 1.19}$ \\

&
& QF-Attack
& $65.31 \textcolor{gray}{\scriptstyle \pm 1.69}$
& $83.19 \textcolor{gray}{\scriptstyle \pm 1.08}$
& $38.45 \textcolor{gray}{\scriptstyle \pm 2.11}$
& $52.14 \textcolor{gray}{\scriptstyle \pm 1.87}$ \\

&
& SneakyPrompt
& $53.47 \textcolor{gray}{\scriptstyle \pm 1.78}$
& $52.94 \textcolor{gray}{\scriptstyle \pm 1.84}$
& $94.26 \textcolor{gray}{\scriptstyle \pm 0.82}$
& $67.12 \textcolor{gray}{\scriptstyle \pm 1.49}$ \\

\cmidrule(lr){2-7}

& \multirow{3}{*}{Untargeted}
& ATM
& $54.38 \textcolor{gray}{\scriptstyle \pm 1.85}$
& $53.97 \textcolor{gray}{\scriptstyle \pm 1.91}$
& $62.16 \textcolor{gray}{\scriptstyle \pm 1.62}$
& $57.21 \textcolor{gray}{\scriptstyle \pm 1.73}$ \\

&
& QF-Attack
& $74.51 \textcolor{gray}{\scriptstyle \pm 1.38}$
& $73.26 \textcolor{gray}{\scriptstyle \pm 1.44}$
& $75.92 \textcolor{gray}{\scriptstyle \pm 1.27}$
& $75.18 \textcolor{gray}{\scriptstyle \pm 1.31}$ \\

&
& VerifyPrompt (Ours)
& $\mathbf{98.17} \textcolor{gray}{\scriptstyle \pm 0.52}$
& $\mathbf{96.30} \textcolor{gray}{\scriptstyle \pm 0.58}$
& $\mathbf{96.12} \textcolor{gray}{\scriptstyle \pm 0.63}$
& $\mathbf{98.45} \textcolor{gray}{\scriptstyle \pm 0.50}$ \\

\midrule

\multirow{6}{*}{DALL-E 3}
& \multirow{3}{*}{Targeted}
& RIATIG
& $75.29 \textcolor{gray}{\scriptstyle \pm 1.27}$
& $72.81 \textcolor{gray}{\scriptstyle \pm 1.39}$
& $81.56 \textcolor{gray}{\scriptstyle \pm 1.13}$
& $77.03 \textcolor{gray}{\scriptstyle \pm 1.22}$ \\

&
& QF-Attack
& $62.71 \textcolor{gray}{\scriptstyle \pm 1.81}$
& $67.19 \textcolor{gray}{\scriptstyle \pm 1.61}$
& $48.36 \textcolor{gray}{\scriptstyle \pm 1.94}$
& $56.14 \textcolor{gray}{\scriptstyle \pm 1.78}$ \\

&
& SneakyPrompt
& $55.38 \textcolor{gray}{\scriptstyle \pm 1.74}$
& $53.18 \textcolor{gray}{\scriptstyle \pm 1.86}$
& $85.74 \textcolor{gray}{\scriptstyle \pm 1.09}$
& $66.19 \textcolor{gray}{\scriptstyle \pm 1.53}$ \\

\cmidrule(lr){2-7}

& \multirow{3}{*}{Untargeted}
& ATM
& $63.14 \textcolor{gray}{\scriptstyle \pm 1.72}$
& $61.89 \textcolor{gray}{\scriptstyle \pm 1.78}$
& $71.72 \textcolor{gray}{\scriptstyle \pm 1.46}$
& $66.42 \textcolor{gray}{\scriptstyle \pm 1.59}$ \\

&
& QF-Attack
& $69.35 \textcolor{gray}{\scriptstyle \pm 1.48}$
& $65.18 \textcolor{gray}{\scriptstyle \pm 1.57}$
& $81.74 \textcolor{gray}{\scriptstyle \pm 1.21}$
& $63.18 \textcolor{gray}{\scriptstyle \pm 1.66}$ \\

&
& VerifyPrompt (Ours)
& $\mathbf{92.42} \textcolor{gray}{\scriptstyle \pm 0.71}$
& $\mathbf{91.17} \textcolor{gray}{\scriptstyle \pm 0.76}$
& $\mathbf{93.91} \textcolor{gray}{\scriptstyle \pm 0.64}$
& $\mathbf{92.01} \textcolor{gray}{\scriptstyle \pm 0.69}$ \\

\bottomrule
\end{tabular}
\end{adjustbox}
\end{table*}

\textbf{Updating the Policy Network.}  
After each interaction, we update the policy network parameters using the REINFORCE algorithm~\cite{williams1992simple}:
\begin{equation}
\theta \leftarrow \theta + \alpha \nabla_\theta \log \pi_\theta(a_t|s_t) \mathcal{R}_t,
\end{equation}
where \(\alpha\) is the learning rate. This update encourages actions that lead to greater reductions in the total loss. The search continues until the query budget is exhausted or convergence is achieved, and the best-performing suffix is returned as the final verify prompt.

\subsection{Threshold detection strategy}

We adopt the 3-Sigma rule to define a threshold for distinguishing CLIP-text scores based on their statistical properties. Instead of relying only on the average score, which ignores score variance and may lead to misclassification, we incorporate the score variance into the threshold calculation. Let the CLIP-text scores produced by the reference model be denoted by
\begin{equation}
X_r \sim \mathcal{N}(\mu_r,\sigma_r^2),
\end{equation}
where $\mu_r$ and $\sigma_r$ are the mean and standard deviation, respectively. The consistency interval of the reference model is defined as
\begin{equation}
\mathcal{I}_r
\left[
\mu_r-3\sigma_r,
\mu_r+3\sigma_r
\right].
\end{equation}
Under the normality assumption, approximately $99.7\%$ of the CLIP-text scores generated by the reference model are expected to fall within this interval.

Based on this interval, we further formulate model consistency determination as an equivalence hypothesis test. Let $\mu_t$ denote the expected CLIP-text score of the target model, and let the equivalence margin be defined as
\begin{equation}
\Delta = 3\sigma_r.
\end{equation}
The hypotheses are formulated as
\begin{equation}
\begin{aligned}
H_0 &: \left|\mu_t-\mu_r\right| \geq \Delta, \
H_1 &: \left|\mu_t-\mu_r\right| < \Delta,
\end{aligned}
\end{equation}
where $H_0$ indicates that the target model is inconsistent with the reference model, whereas $H_1$ indicates that the two models are statistically equivalent within the predefined tolerance.

Given $n$ CLIP-text scores obtained from the target model, we calculate their sample mean $\bar{X}_t$. The target model is regarded as consistent with the reference model when
\begin{equation}
\mu_r-3\sigma_r
<
\bar{X}_t
<
\mu_r+3\sigma_r,
\end{equation}
which is equivalent to
\begin{equation}
\left|\bar{X}_t-\mu_r\right|<\Delta.
\end{equation}
Otherwise, the null hypothesis is retained, and the target model is regarded as inconsistent with the reference model. In this way, the proposed hypothesis-testing formulation does not change the original 3-Sigma-based detection strategy. Still, it provides a statistically interpretable criterion for determining whether the target model is consistent with the reference model.

\begin{figure*}[]
\centerline{\includegraphics[width=0.8\linewidth]{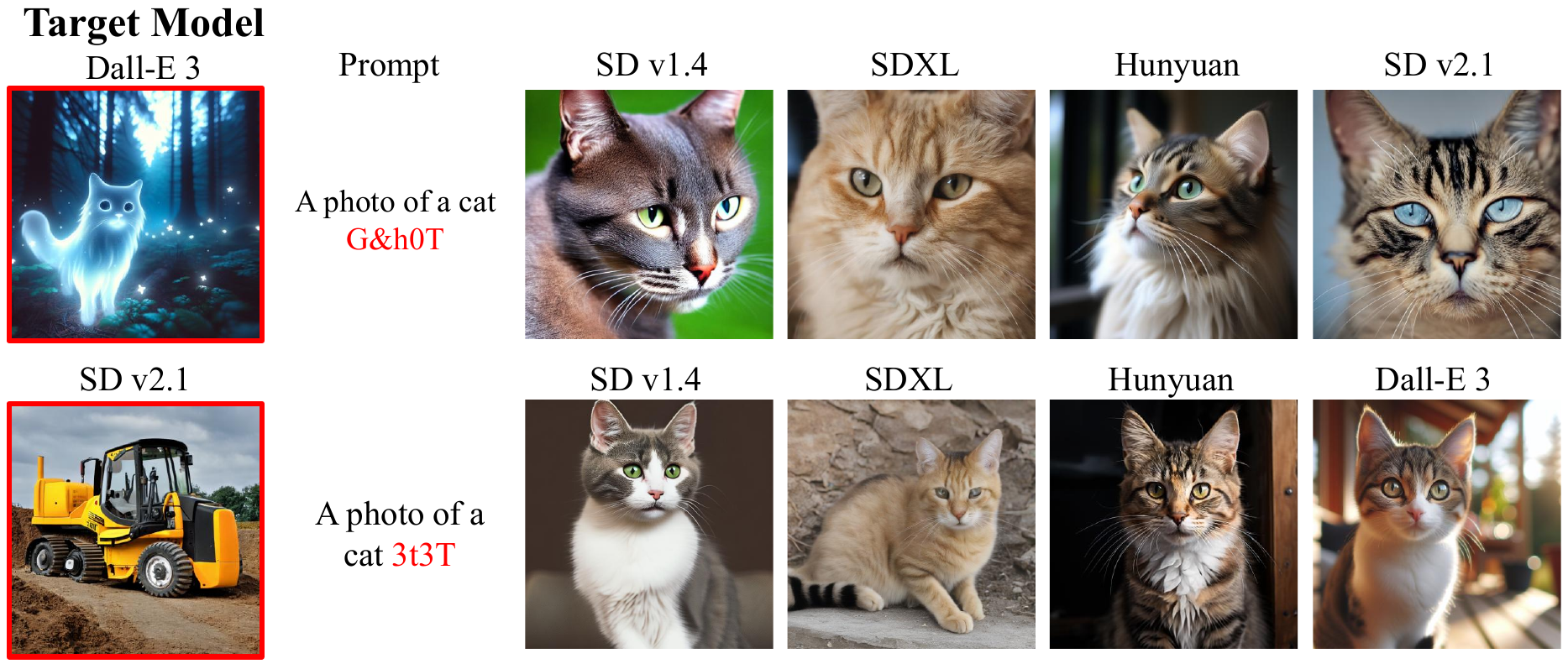}}
\caption{Generated images of verify prompts under different T2I models}
\label{Generated images of verify prompts under different T2I models}
\end{figure*}

\section{Evaluation}

\subsection{Experimental Settings}


\textbf{Dataset.} We use GPT-4o to generate 10 basic prompts as templates for our experiments (see appendix), as in practice verifiers can freely choose their own original prompts during verification.

\textbf{Model Architecture.} Without loss of generality, we selected four representative models, including three open-source models and one closed-source model (Stable Diffusion v1.4~\cite{stable_diffusion_v1_4}, Stable Diffusion v2.1~\cite{stable_diffusion_v2_1}, Hunyuan DiT v1.2~\cite{Hunyuan_DiT_v1.2}, and DALL-E 3~\cite{dall-e-3}) for verification. To better reflect real-world scenarios, we also included 6 unknown models as other models (SDXL~\cite{sdxl}, FLUX.1~\cite{flux_1}, Playground-v2.5~\cite{playground_v2_5}, Playground-v2~\cite{playground_v2}, DALL-mini~\cite{dalle_mini}, Stable Diffusion v3~\cite{stable_diffusion_v3}).


\textbf{Evaluation Metrics.}
We use four commonly adopted metrics, namely Accuracy, Precision, Recall, and F1-Score, to evaluate the performance of model verification. Since T2I models exhibit inherent generation randomness, we conduct each experiment 3 times and report these metrics as the mean $\pm$ standard deviation.

\textbf{Configurations of VerifyPrompt.}
For the VerifyPrompt-RL optimization of VerifyPrompt, we set the prompt to be sampled 5 times, with a total of 300 queries and 5 appended perturbation characters. We also evaluate the impact of different query numbers and the number of perturbation characters in the subsequent experiments. The weights of the loss function are set as \(\lambda_1 = 0.8\) and \(\lambda_2 = 0.2\).

\textbf{Baseline Selection.} We selected three types of targeted adversarial attacks (RIATIG~\cite{liu2023riatig}, QF-Attack~\cite{zhuang2023pilot}, SneakyPrompt~\cite{yang2024sneakyprompt}) and two types of untargeted adversarial attacks (ATM~\cite{ATM}, QF-Attack~\cite{zhuang2023pilot}) to generate adversarial prompts, which are subsequently used for model verification. For a fair comparison, we determine a CLIP-text score threshold for each adversarial attack method.

\textbf{Evaluation Details.}
For each model to be verified, we set a binary ground truth label indicating whether it is the target model (positive, label~1) or not (negative, label~0). We use the verify prompt to generate images for each model and classify their label based on CLIP-text thresholds. If the predicted label matches the ground truth label, we consider the verification successful. The final metrics were averaged across all samples (including labels~1 and 0). For each original prompt, we generated 30 verify prompts and computed their average score for evaluation.

\subsection{Verification Performance Evaluation}
\textbf{Effectiveness of VerifyPrompt.}
Table~\ref{table1} shows the performance of model verification using our verify prompts and other adversarial prompts.
As presented in Table~\ref{table1}, VerifyPrompt achieves the best results in all four metrics, accuracy, precision, recall, and F1-Score, in all four target models.
This is because VerifyPrompt incorporates a reduced transferability objective function, whereas other adversarial attack methods do not specifically enhance reduced transferability.
As a result, the prompts generated for the target models can be successfully transferred to other models, reducing the effectiveness of model verification.
This indicates that the adversarial prompts generated by VerifyPrompt are more effective for model verification than those generated by other adversarial attack methods. This is because VerifyPrompt incorporates an objective function designed to reduce transferability, whereas adversarial prompts generated by other adversarial attack methods are inevitably transferable to some extent.

\textbf{Visualization Results.} Fig.~\ref{Generated images of verify prompts under different T2I models} presents a set of examples generated by our VerifyPrompt. For the original prompts "A photo of a cat," models correctly generate images of a cat, respectively. However, when using the verify prompts generated by VerifyPrompt, the target model produces an image of an excavator. In contrast, the other models still generate images that correspond to the original prompts. This demonstrates the effectiveness of our method in targeting specific models while preserving the semantics of the original prompt in others.

\begin{figure*}[t]
    \centering
    \begin{subfigure}[b]{0.23\linewidth}
        \centering
        \includegraphics[width=\linewidth]{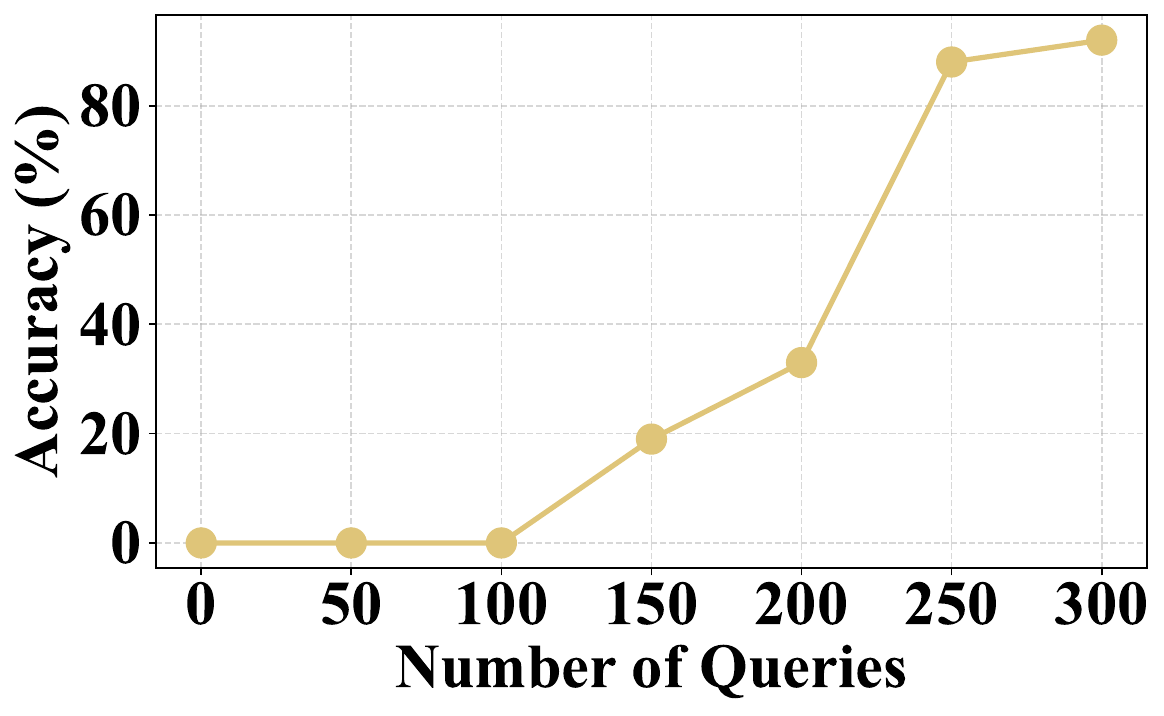}
        \caption{Stable Diffusion v1.4}
        \label{fig:sub1}
    \end{subfigure}
    \hfill
    \begin{subfigure}[b]{0.23\linewidth}
        \centering
        \includegraphics[width=\linewidth]{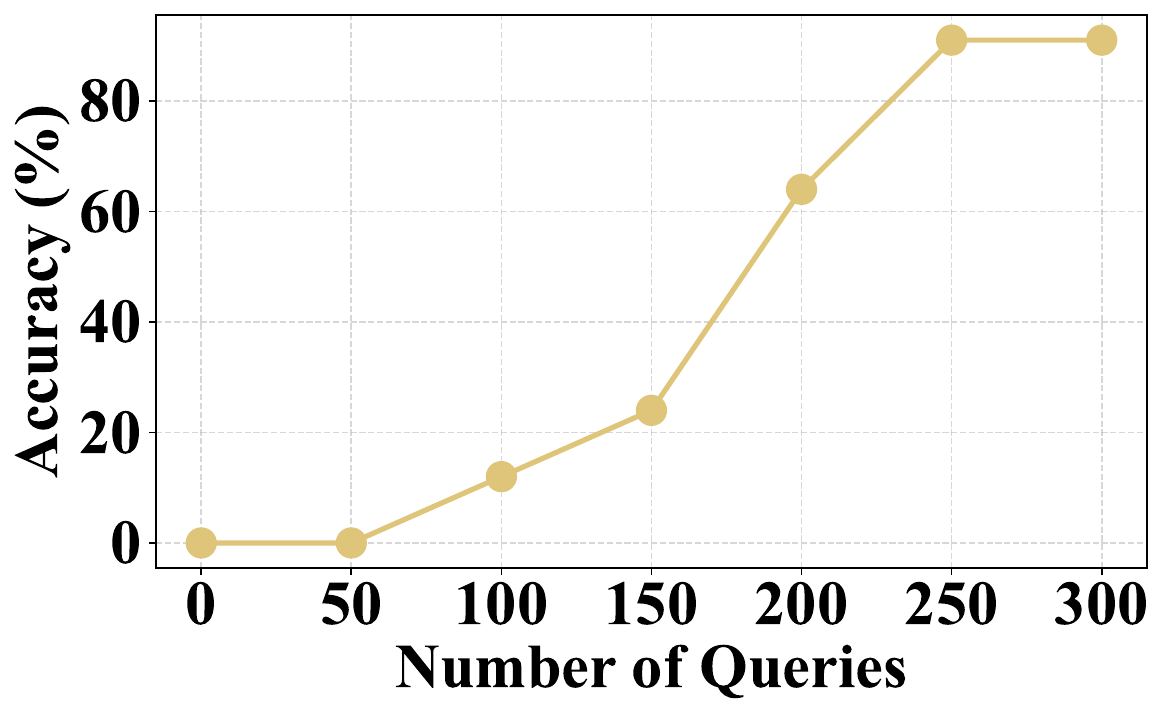}
        \caption{Stable Diffusion v2.1}
        \label{fig:sub2}
    \end{subfigure}
    \hfill
        \begin{subfigure}[b]{0.23\linewidth}
        \centering
        \includegraphics[width=\linewidth]{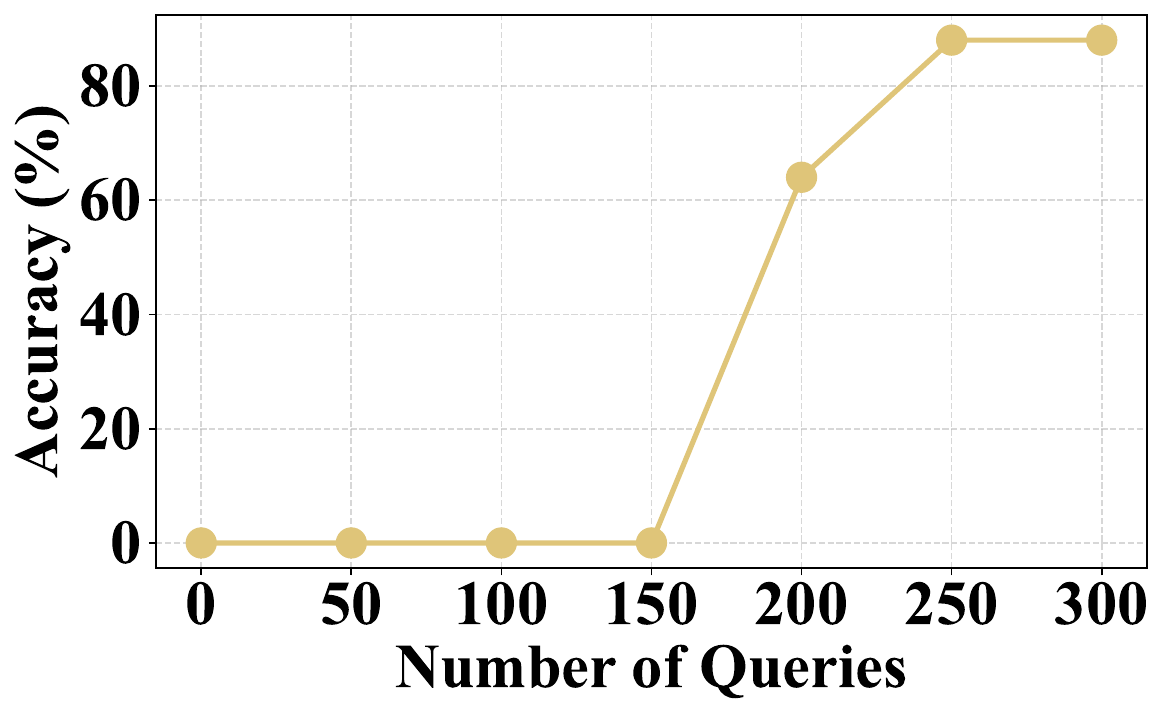}
        \caption{Hunyuan DiT v1.2}
        \label{fig:sub3}
    \end{subfigure}
    \hfill
        \begin{subfigure}[b]{0.23\linewidth}
        \centering
        \includegraphics[width=\linewidth]{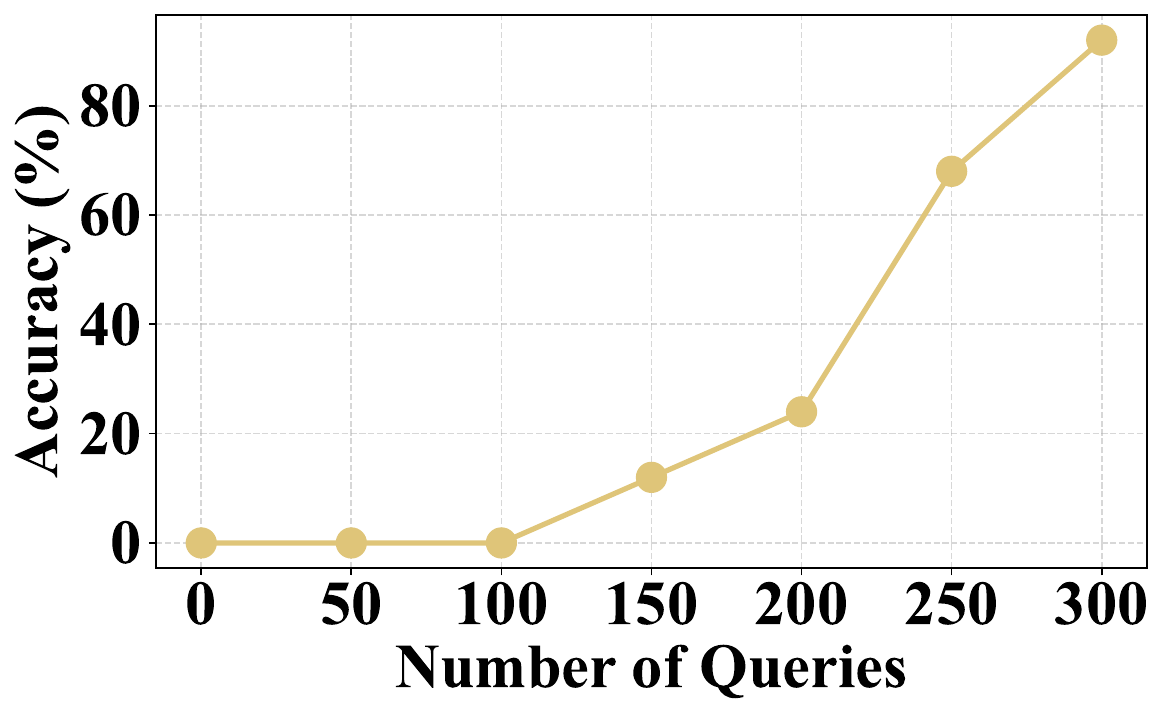}
        \caption{DALL-E 3}
        \label{fig:sub4}
    \end{subfigure}
    \caption{Impact of query budget}
    \label{Impact of Query Budget}
\end{figure*}

\begin{table}[t]
\centering
\caption{Ablation study of different objective functions on four models}
\begin{adjustbox}{max width=\linewidth}
\begin{tabular}{lccccc}
\toprule
\textbf{Model} & \textbf{Objective Function} & \textbf{Accuracy} & \textbf{Precision} & \textbf{Recall} & \textbf{F1-Score} \\
\midrule
\multirow{2}{*}{Stable Diffusion v1.4} 
  & \( f_1 \)           & 56.35 & 88.42 & 48.10 & 62.13 \\
  & \( f_1+f_2 \)       & 92.50 & 91.36 & 92.14 & 91.75 \\
\midrule
\multirow{2}{*}{Stable Diffusion v2.1} 
  & \( f_1 \)           & 71.42 & 89.25 & 60.83 & 72.71 \\
  & \( f_1+f_2 \)       & 90.13 & 92.47 & 89.96 & 91.20 \\
\midrule
\multirow{2}{*}{Hunyuan DiT v1.2} 
  & \( f_1 \)           & 68.90 & 88.97 & 57.45 & 69.92 \\
  & \( f_1+f_2 \)       & 88.76 & 90.82 & 88.31 & 89.55 \\
\midrule
\multirow{2}{*}{DALL\hbox{-}E 3} 
  & \( f_1 \)           & 74.21 & 89.64 & 65.48 & 75.62 \\
  & \( f_1+f_2 \)       & 91.87 & 93.12 & 91.42 & 92.26 \\
\bottomrule
\end{tabular}
\end{adjustbox}
\label{tab:ablation_multi_model}
\end{table}

\begin{figure}[ht]
    \centering
    \begin{subfigure}[b]{0.45\linewidth}
        \centering
        \includegraphics[width=\linewidth]{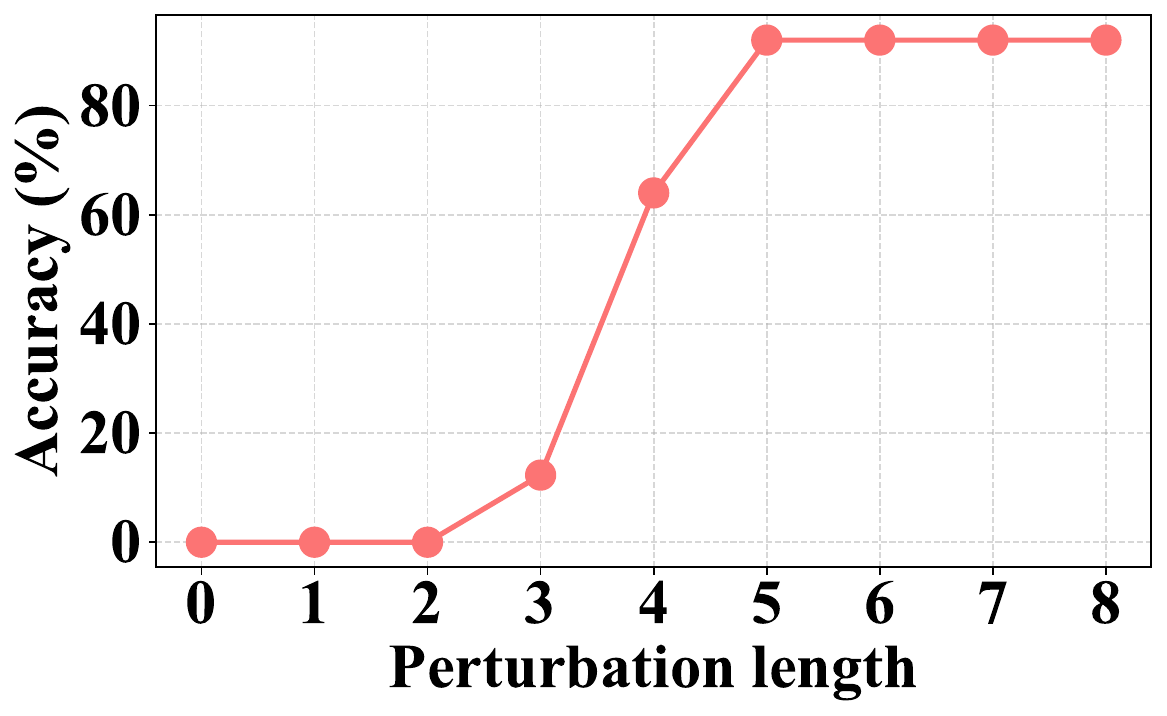}
        \caption{Stable Diffusion v1.4}
        \label{fig:sub1}
    \end{subfigure}
    \hfill
    \begin{subfigure}[b]{0.45\linewidth}
        \centering
        \includegraphics[width=\linewidth]{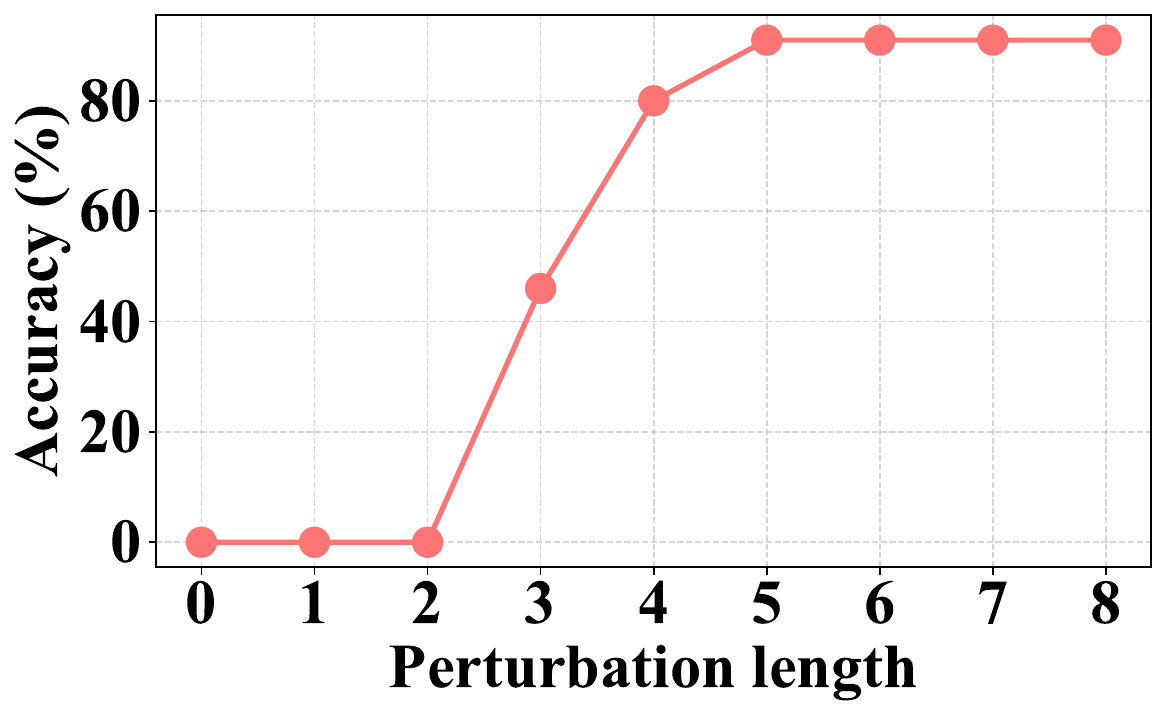}
        \caption{Stable Diffusion v2.1}
        \label{fig:sub2}
    \end{subfigure}
    \hfill
        \begin{subfigure}[b]{0.45\linewidth}
        \centering
        \includegraphics[width=\linewidth]{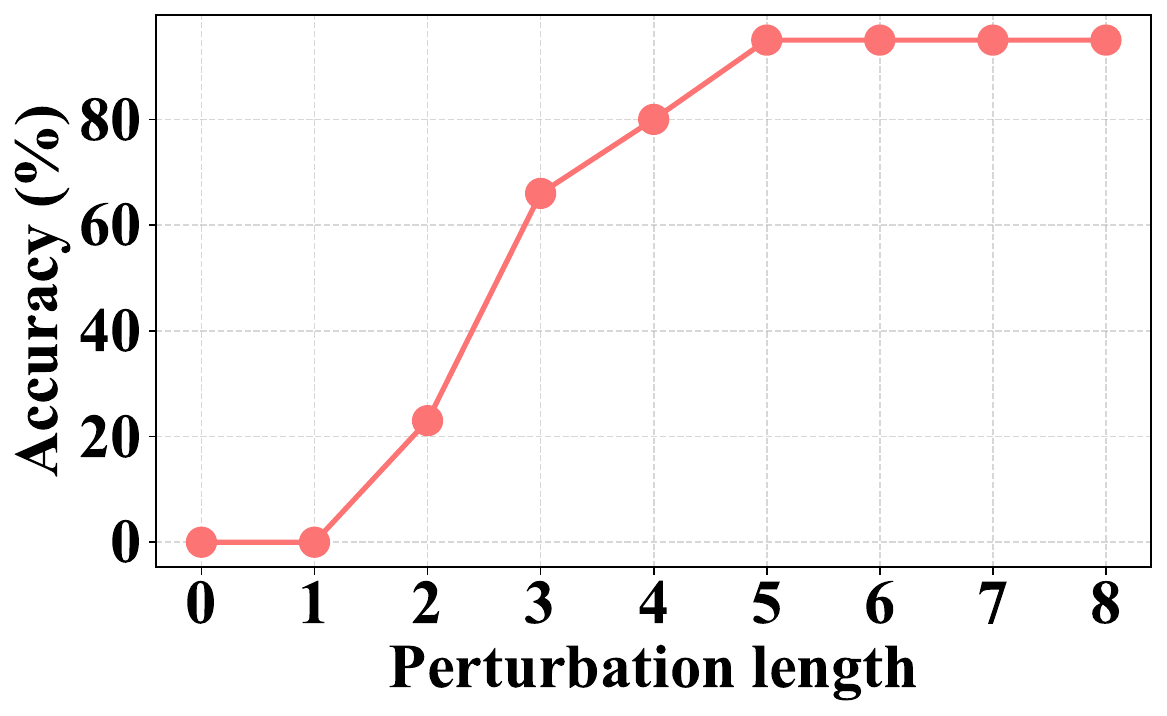}
        \caption{Hunyuan DiT v1.2}
        \label{fig:sub3}
    \end{subfigure}
    \hfill
        \begin{subfigure}[b]{0.45\linewidth}
        \centering
        \includegraphics[width=\linewidth]{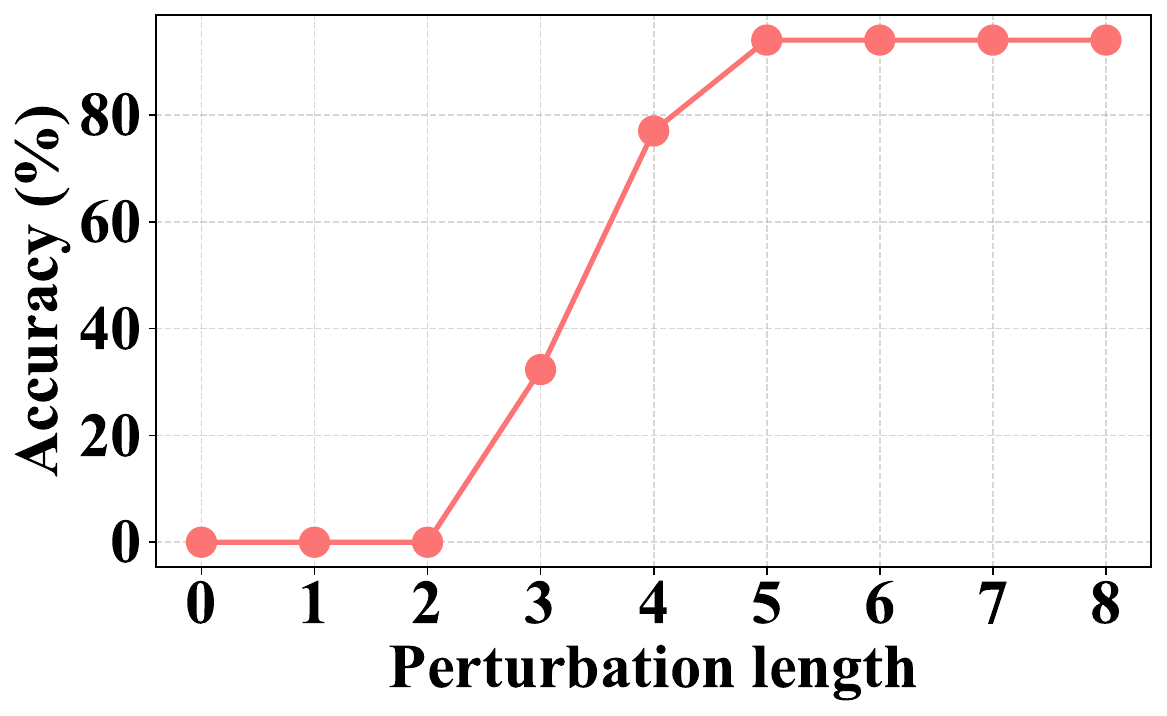}
        \caption{DALL-E 3}
        \label{fig:sub4}
    \end{subfigure}
    \caption{Impact of prompt perturbation length}
    \label{mpact of prompt perturbation length}
\end{figure}

\begin{table}[t]
\centering
\caption{Comparison of different optimization methods on four models.}
\label{tab:optimization_methods_multi_model}
\begin{adjustbox}{max width=\linewidth}
\begin{tabular}{lcccc}  
\toprule
\textbf{Model} & \textbf{Optimization Method} & \textbf{Accuracy} & \textbf{Number of Queries} \\
\midrule
\multirow{5}{*}{Stable Diffusion v1.4} 
& Greedy            & 50.23 & 1384 \\
& NSGA-II           & 80.74 & 721  \\
& GA                & 81.21 & 1047 \\
& PSO                  & 73.24 & 892  \\
& VerifyPrompt-RL   & \textbf{90.08} & \textbf{122} \\
\midrule
\multirow{5}{*}{Stable Diffusion v2.1} 
& Greedy            & 32.14 & 1298 \\
& NSGA-II           & 79.65 & 759  \\
& GA                & 67.34 & 987  \\
& PSO               & 70.87 & 902  \\
& VerifyPrompt-RL   & \textbf{88.90} & \textbf{140} \\
\midrule
\multirow{5}{*}{Hunyuan DiT v1.2} 
& Greedy            & 22.45 & 1415 \\
& NSGA-II           & 78.80 & 732  \\
& GA                & 66.10 & 1112 \\
& PSO               & 71.52 & 919  \\
& VerifyPrompt-RL   & \textbf{87.63} & \textbf{186} \\
\midrule
\multirow{5}{*}{DALL-E 3} 
& Greedy            & 13.08 & 1852 \\
& NSGA-II           & 68.96 & 1781  \\
& GA                & 29.43 & 1562 \\
& PSO               & 34.66 & 1687  \\
& VerifyPrompt-RL   & \textbf{89.45} & \textbf{283} \\
\bottomrule
\end{tabular}
\end{adjustbox}
\end{table}

\subsection{Computational Overhead}
We evaluate the effect of different query budgets on model verification accuracy using four different models. For all experiments, we append a 5-character suffix as the adversarial prompt. As shown in Fig.~\ref{Impact of Query Budget}, once the number of queries exceeds 250, the verification accuracy consistently surpasses 90\%. This demonstrates that VerifyPrompt-RL can efficiently discover effective verify prompts with a limited number of queries. Moreover, the computational cost remains within an acceptable range for users, further highlighting the practicality of our approach.

\subsection{Ablation study}

\textbf{Objective Function.}
We evaluate the effect on the objective function in Eq.\ref{eq:total_loss} to further demonstrate the contribution of enhancing the interaction of adversarial perturbations. For each prompt, we generate 30 adversarial prompts and take their average value as the final result. As shown in Table~\ref{tab:ablation_multi_model}, after incorporating
\( f_2 \), the verification accuracy improves significantly. This further confirms that strengthening the interaction of adversarial perturbations effectively reduces attack transferability, whereas using standard adversarial prompts alone leads to lower verification accuracy.

\textbf{Prompt Perturbation Length.} We analyzed the impact of different perturbation lengths on model verification accuracy. We selected the prompt "A photo of a cat" as the original prompt.
Fig.~\ref{mpact of prompt perturbation length} shows the accuracy for perturbation lengths ranging from 0 to 8. We found that accuracy gradually increases before a perturbation length of 5 and remains nearly unchanged afterward. This indicates that a perturbation length of 5 is sufficient for verification with this prompt, as longer perturbations do not improve accuracy. Therefore, we set the perturbation length to 5 to ensure accuracy.

\textbf{Optimization Method.} We evaluate the model verification performance of different optimization methods under a fixed query budget of 300 and report the average number of queries required to generate a verify prompt. For comparison, we include four representative black-box optimization algorithms: Greedy~\cite{greedy}, NSGA-II~\cite{NSGA-II}, Genetic Algorithm (GA)~\cite{ga}, and Particle Swarm Optimization (PSO)~\cite{pso}. For each method, we generate 30 verify prompts and report the average verification accuracy and number of queries required. 
As shown in Table~\ref{tab:optimization_methods_multi_model}, under the same query budget of 300, our VerifyPrompt-RL consistently outperforms other black-box optimization methods. Meanwhile, VerifyPrompt-RL requires significantly fewer average queries than the other methods. This is because the RL-based search strategy explores a larger solution space and is less likely to get trapped in local minima, which often lead to optimization failure. In addition, our group-based reward design ensures more stable and reliable optimization performance.

\begin{table}[t]
    \centering
    \caption{Model verification performance of VerifyPrompt on Hugging Face APIs.}
    \label{tab:huggingface_verification}
    \resizebox{\linewidth}{!}{
    \begin{tabular}{llcc}
        \toprule
        \textbf{Reference Model} & \textbf{Target Model} & \textbf{R-1} & \textbf{R-3} \\
        \midrule
        \multirow{7}{*}{Stable Diffusion v2.1}
        & Stable Diffusion v1.4~\cite{stable_diffusion_v1_4} & 0/10 & 1/10 \\
        & Hunyuan DiT v1.2~\cite{Hunyuan_DiT_v1.2} & 0/10 & 0/10 \\
        & Playground-v2.5~\cite{playground_v2_5} & 0/10 & 0/10 \\
        & Stable Diffusion v3~\cite{stable_diffusion_v3} & 0/10 & 0/10 \\
        & Hyper-SD~\cite{Hyper-SD} & 0/10 & 0/10 \\
        & SDXL~\cite{sdxl} & 0/10 & 0/10 \\
        & FLUX.1~\cite{flux_1} & 0/10 & 0/10 \\
        & DALL-mini~\cite{dalle_mini} & 0/10 & 0/10 \\
        \rowcolor{gray!15}
        & Stable Diffusion v2.1~\cite{stable_diffusion_v2_1} & \textbf{10/10} & \textbf{10/10} \\
        \midrule
        \multirow{7}{*}{Hunyuan DiT v1.2}
        & Stable Diffusion v1.4~\cite{stable_diffusion_v1_4} & 0/10 & 0/10 \\
        & Stable Diffusion v2.1~\cite{stable_diffusion_v2_1} & 0/10 & 0/10 \\
        & Playground-v2.5~\cite{playground_v2_5} & 0/10 & 0/10 \\
        & Stable Diffusion v3~\cite{stable_diffusion_v3} & 0/10 & 0/10 \\
        & SDXL~\cite{sdxl} & 0/10 & 0/10 \\
        & FLUX.1~\cite{flux_1} & 0/10 & 0/10 \\
        & DALL-mini~\cite{dalle_mini} & 0/10 & 0/10 \\
        \rowcolor{gray!15}
        & Hunyuan DiT v1.2~\cite{Hunyuan_DiT_v1.2} & \textbf{10/10} & \textbf{10/10} \\
        \bottomrule
    \end{tabular}
    }
\end{table}

\section{Case study}

In this section, we employ VerifyPrompt to validate whether third-party platforms host models consistent with their claimed identities. We conduct verification on Stable Diffusion v2.1 and Hunyuan DiT v1.2 deployed on Hugging Face, and DALL·E 3 deployed via One-API. The reference models are the official checkpoints released on GitHub and the OpenAI API, while the target models are accessed through Hugging Face and One-API endpoints. Since the ground-truth identities of the target models are not publicly available, we examine the top-R ranked verification results and consider a match successful only when the retrieved images from the reference model deviate significantly from the semantics of the original prompt, indicating that the target model behaves consistently with its claimed identity.

\begin{figure}[t]
\centerline{\includegraphics[width=\linewidth]{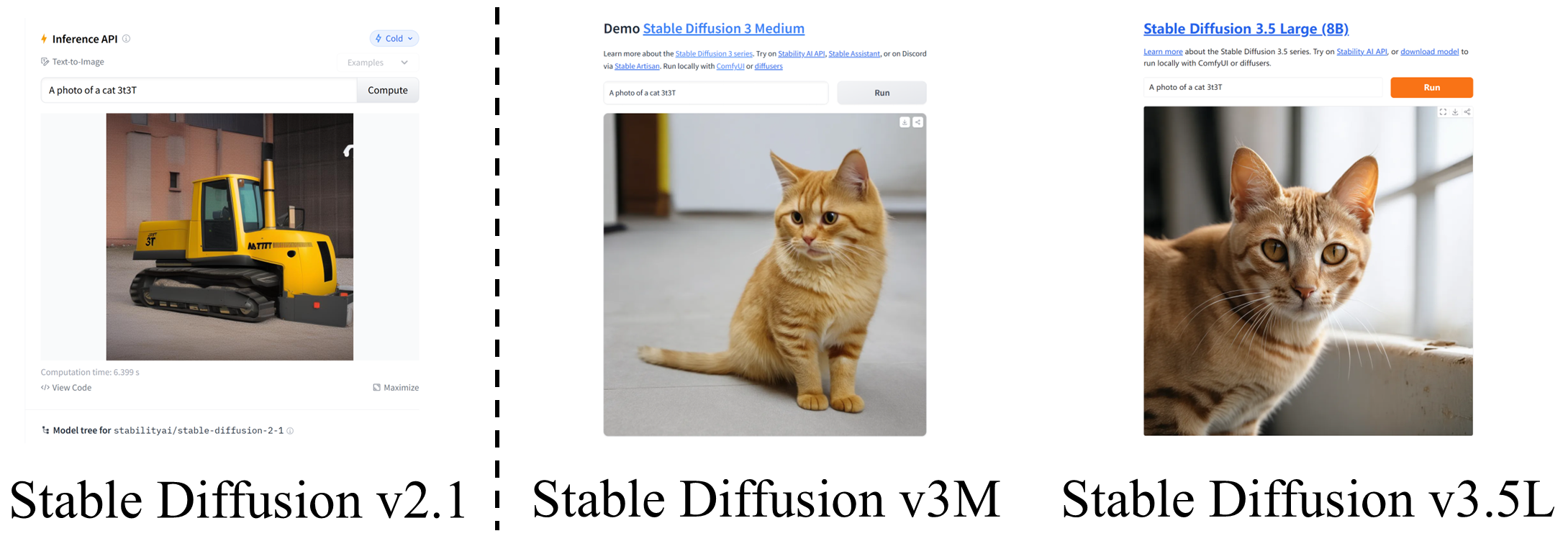}}
\caption{Visualization results of our verify prompt for model APIs from Hugging Face}
\label{Visualization results of our verify prompt for model APIs
from Hugging Face}
\end{figure}

\begin{figure}[t]
    \centering
    \begin{subfigure}[b]{0.4\linewidth}
        \centering
        \includegraphics[width=\linewidth]{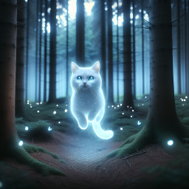}
        \caption{OpenAI}
        \label{fig:sub1}
    \end{subfigure}
    \hfill
    \begin{subfigure}[b]{0.4\linewidth}
        \centering
        \includegraphics[width=\linewidth]{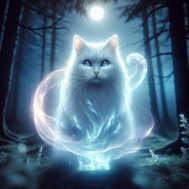}
        \caption{One-API}
        \label{fig:sub2}
    \end{subfigure}
    \caption{Visualization results of our verify prompt of DALL-E 3 from different APIs}
    \label{One-api}
\end{figure}
\subsection{Hugging Face}
\textbf{Main Results.}
We use verify prompts generated from Stable Diffusion v2.1 and Hunyuan DiT v1.2 to evaluate various model APIs. As shown in Table~\ref{tab:huggingface_verification}, only the corresponding target APIs yield high R-1 and R-3 scores, while all other models achieve near-zero values. This demonstrates two key points: (1) the API-hosted models for Stable Diffusion v2.1 and Hunyuan DiT v1.2 are consistent with their claimed identities; and (2) other APIs do not appear to be repackaged or misrepresentations of these models.

\textbf{Visualization Results.}
We further provide a visual comparison of the images generated by the reference models and the different API-hosted models. As shown in Fig.~\ref{Visualization results of our verify prompt for model APIs
from Hugging Face}, only the APIs that claim to use the target model produce images that are consistent with those generated by the corresponding reference models. In contrast, all other APIs generate images with different semantics. This further confirms that the models behind the Stable Diffusion v2.1 and Hunyuan DiT v1.2 APIs are indeed consistent with their claimed identities, and that other APIs do not appear to misuse or repackage these models.

\begin{table}[t]
    \centering
    \caption{Model verification performance of VerifyPrompt on One-API APIs.}
    \label{tab:One-API}
    \resizebox{0.8\linewidth}{!}{
    \begin{tabular}{llcc}
        \toprule
        \textbf{Reference Model} & \textbf{Target Model} & \textbf{R-1} & \textbf{R-3} \\
        \midrule
        \multirow{1}{*}{DALL-E 3~\cite{dall-e-3}}
        & DALL-E 3& 9/10 & 10/10 \\
        \bottomrule
    \end{tabular}
    }
\end{table}

\subsection{One-API}

\begin{figure}[t]
    \centering
    \begin{subfigure}[b]{0.31\linewidth}
        \centering
        \includegraphics[width=\linewidth]{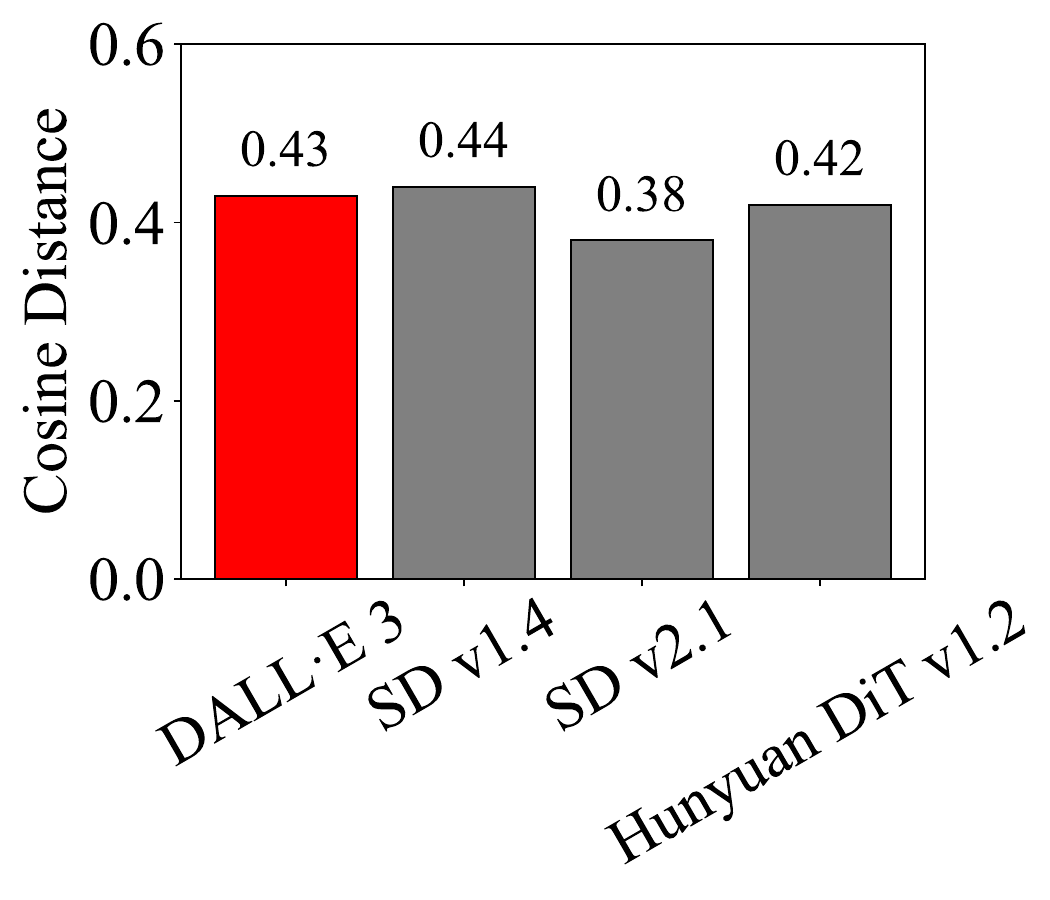}
        \caption{RIATIG}
        \label{fig:sub1}
    \end{subfigure}
    \hfill
    \begin{subfigure}[b]{0.31\linewidth}
        \centering
        \includegraphics[width=\linewidth]{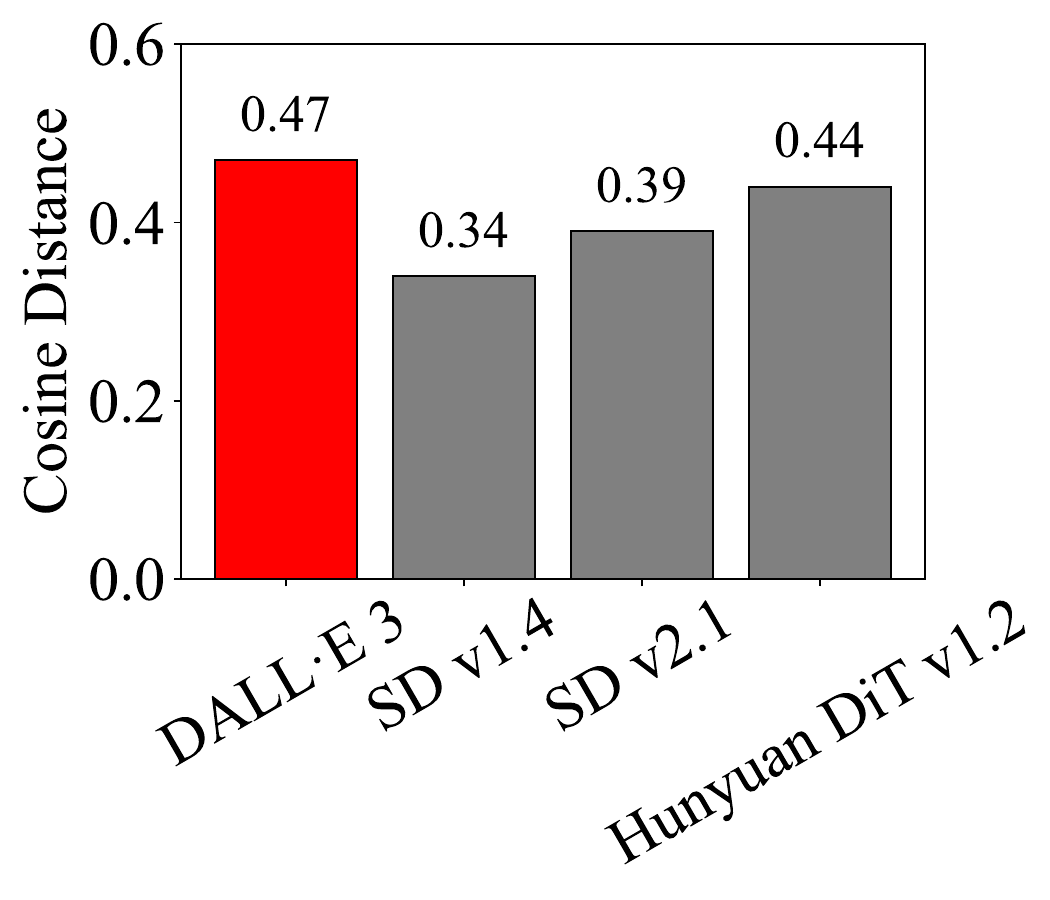}
        \caption{SneakyPrompt}
        \label{fig:sub2}
    \end{subfigure}
    \hfill
    \begin{subfigure}[b]{0.31\linewidth}
        \centering
        \includegraphics[width=\linewidth]{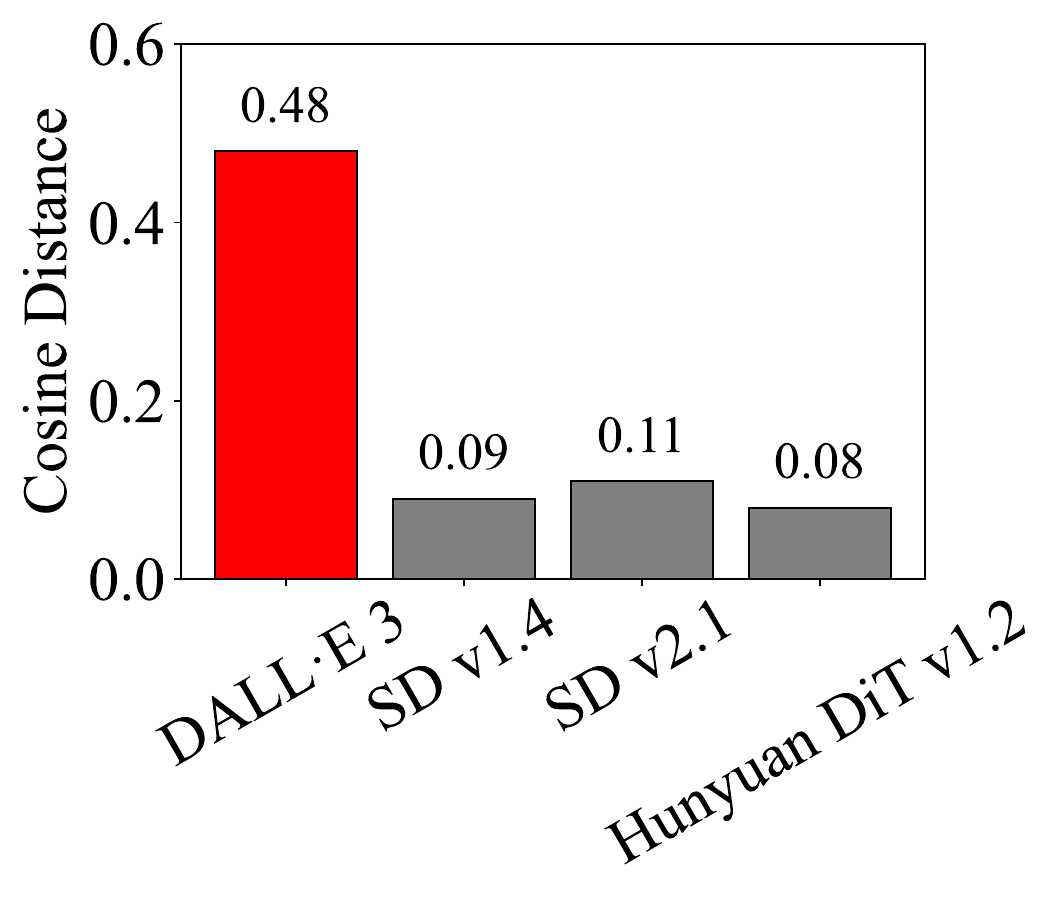}
        \caption{VerifyPrompt}
        \label{fig:sub2}
    \end{subfigure}
    \caption{Semantic shift between original and adversarial prompts using DALL-E 3 as reference model.}
    \label{Semantic Shift}
\end{figure}

\begin{table}[t]
\centering
\caption{Verification accuracy at different fine-tuning steps on Stable Diffusion v2.1}
\label{tab:finetune_accuracy}
\begin{tabular}{c|c}
\toprule
\textbf{Fine-tuning Steps} & \textbf{Verification Accuracy (\%)} \\
\midrule
0       & 0.00 \\
3000    & 72.60 \\
5000    & 86.32 \\
7000    & 88.30 \\
9000    & 85.26 \\
10000   & 93.26    \\
\bottomrule
\end{tabular}
\end{table}

\textbf{Main Results.}
As shown in Table~\ref{tab:One-API}, the images generated by One-API exhibit a high 
R-1 and R-3 scores when compared with those from the official OpenAI API. This strong alignment indicates that the DALL·E 3 APIs provided by One-API are consistent with their claimed model identity.

\textbf{Visualization Results.}
We further provide visualization results of our verify prompts on DALL·E 3 across different APIs. As shown in Fig.~\ref{One-api}, the images generated by OpenAI and One-API are highly similar, further corroborating that the DALL·E 3 model served by One-API is consistent with the official OpenAI deployment.

\section{Discussion}
\textbf{Why does VerifyPrompt work?}
In our approach, we append character-level perturbations at the end of the prompt and strengthen their interaction with the model. This typically leads the perturbations to combine with the original keywords to generate new semantics. In contrast, for other models, such character-level perturbations do not affect the original keywords and are simply interpreted as random, semantically meaningless characters, resulting in failed attacks. Previous adversarial prompt methods (e.g., RIATIG~\cite{liu2023riatig} and SneakyPrompt~\cite{yang2024sneakyprompt}) were originally designed to shift the semantics of the prompt, often by replacing its keywords. Although different models may vary in how they interpret the modified prompt lexically, they tend to remain close in the text embedding space, thereby achieving a transferable adversarial objective (see Fig.~\ref{Semantic Shift}).

\textbf{Impact of fine-tuning.}
We evaluate the impact of fine-tuning on our verification performance. To evaluate the impact of fine-tuning on the effectiveness of VerifyPrompt, we fine-tuned Stable Diffusion v2.1 on a classic text-to-image dataset, Diffusiondb~\cite{wang2022diffusiondb}, with varying numbers of iterations. Specifically, we used a batch size of 32 and set the fine-tuning steps from 3,000 to 10,000. As shown in Table~\ref{tab:finetune_accuracy}, VerifyPrompt is also able to distinguish a fine-tuned model from its original counterpart.
In fact, determining the relationship between a fine-tuned model and its original model is an interesting problem in its own right that can help establish the provenance of the model. Developing effective techniques for model provenance remains an important direction in the future.

\begin{figure}[t]
\centering
\begin{subfigure}[t]{0.48\linewidth}
\centering
\includegraphics[width=\linewidth]{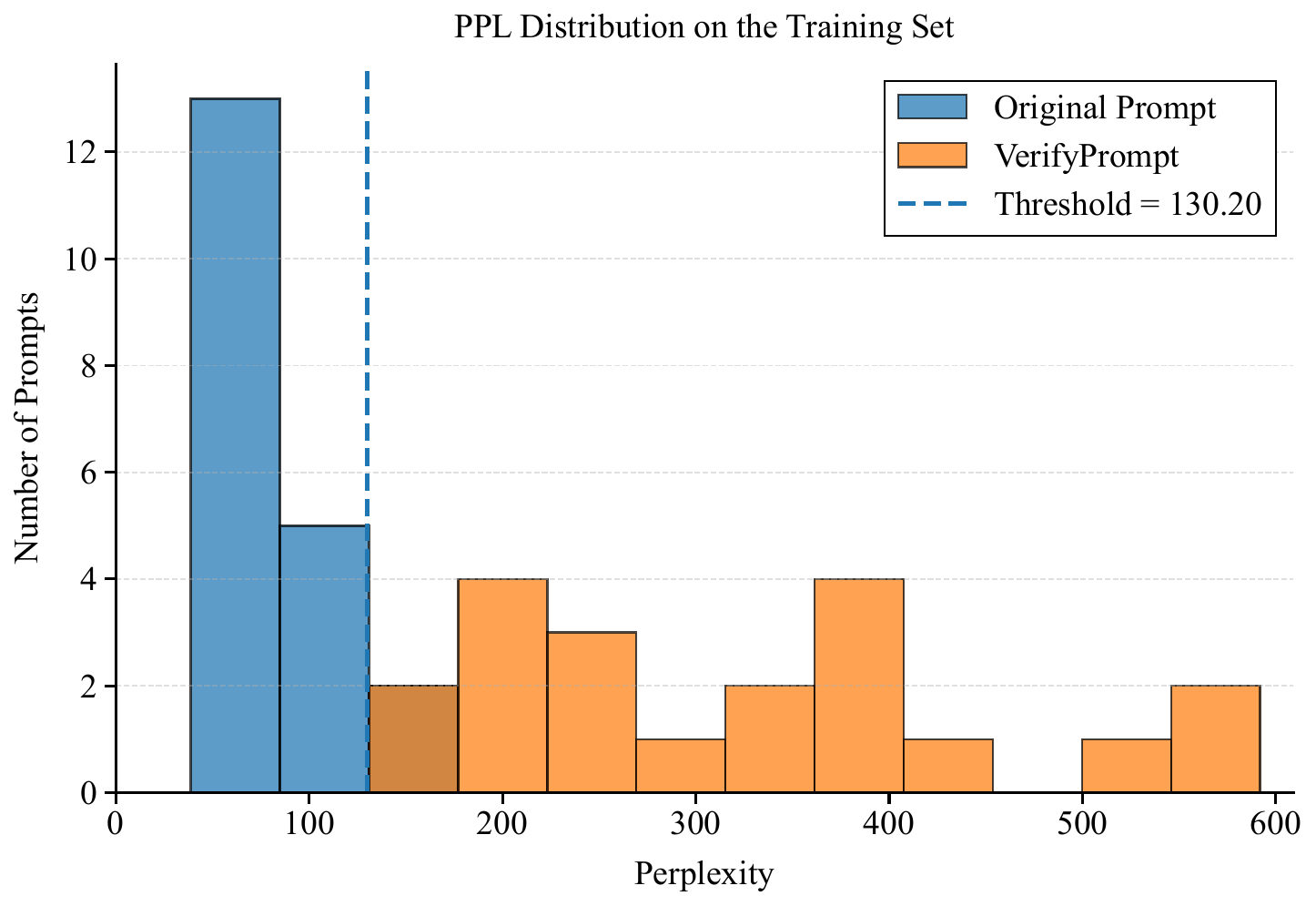}
\caption{Calibration set}
\end{subfigure}
\hfill
\begin{subfigure}[t]{0.48\linewidth}
\centering
\includegraphics[width=\linewidth]{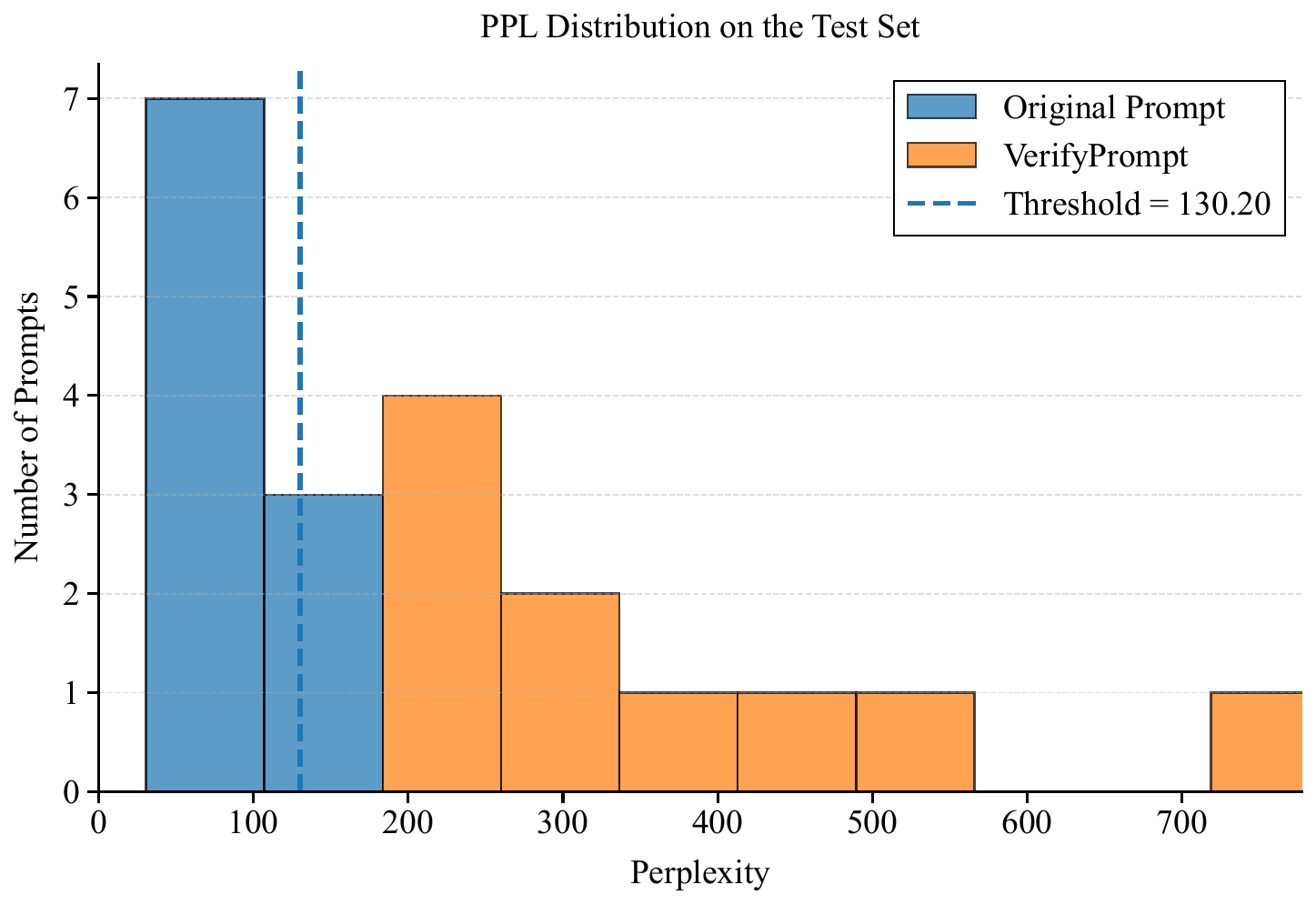}
\caption{Test set}
\end{subfigure}
\caption{PPL distributions of original prompts and VerifyPrompts.}
\label{fig:ppl_distribution}
\end{figure}

\textbf{Limitations of Detection.}
We evaluate whether VerifyPrompt can be detected using prompt perplexity (PPL). Specifically, GPT-2~\cite{radford2019language} is used as the scoring model. We first generate 20 original prompts and VerifyPrompts. The resulting 40 prompts are used as a calibration set, and the detection threshold is selected by maximizing the F1-score. We then independently generate 10 original prompts and 10 VerifyPrompts for testing. A prompt is classified as a VerifyPrompt when its PPL exceeds the calibrated threshold. As shown in Table~\ref{tab:ppl_detection}, the resulting threshold is 130.20, and the detector achieves 95.00\% accuracy, 90.91\% precision, 100.00\% recall, and 95.24\% F1-score. Among the 20 test prompts, all 10 VerifyPrompts are correctly detected, while one original prompt is incorrectly classified.
Fig.~\ref{fig:ppl_distribution} shows that VerifyPrompts generally have higher PPL values than original prompts because random character suffixes are unlikely under the language model and therefore reduce linguistic fluency. Therefore, developing more effective verification prompts that can bypass such detection methods is an important direction for future work. One potential solution is to directly generate semantically meaningful perturbations.

\begin{table}[t]
    \centering
    \caption{Detection performance of the PPL-based detector.}
    \label{tab:ppl_detection}
    \begin{tabular}{cccc}
        \toprule
        Acc. & Prec. & Recall & F1 \\
        \midrule
        95.00 & 90.91 & 100.00 & 95.24 \\
        \bottomrule
    \end{tabular}
\end{table}

\textbf{Adversarial samples without transferability.} Prior studies on adversarial samples have predominantly focused on improving their transferability, with little attention given to reducing it. Indeed, these efforts typically consider the attacker’s perspective, where broader applicability across models is desirable. In contrast, our work is the first to reveal that adversarial prompts with low transferability also hold practical value beyond the commonly studied transferable ones. 
Exploring method to further decrease the transferability of adversarial samples thus represents a promising direction for future research.


\section{Conclusion}

In this paper, we propose VerifyPrompt to verify T2I models behind black-box APIs. VerifyPrompt generates verify prompts that compel the target model to produce a specific image while causing other models to generate entirely different images, thereby achieving model verification. By calculating the CLIP-text score between the images generated by the verify prompt and the original prompt, we set a threshold to distinguish the target model. To enable efficient black-box search for verify prompts, we introduce VerifyPrompt-RL, a reinforcement learning–based strategy that combines multi-sample exploration with intra-group relative error estimation, thereby stabilizing and accelerating the search process.
We evaluate VerifyPrompt across four models, achieving more than 90\% accuracy in model verification. We also utilize VerifyPrompt to verify models on a third-party platform.

\section*{Acknowledgement}
This work is supported by the Sichuan Science and Technology Program under Grant 2025ZNSFSC1490, the National Natural Science Foundation of China under Grant 62402087.

\IEEEpeerreviewmaketitle


{\footnotesize \bibliographystyle{ieeetr}
\bibliography{sample}}

@inproceedings{kim2024wouaf,
  title={Wouaf: Weight modulation for user attribution and fingerprinting in text-to-image diffusion models},
  author={Kim, Changhoon and Min, Kyle and Patel, Maitreya and Cheng, Sheng and Yang, Yezhou},
  booktitle={Proceedings of CVPR},
  pages={8974--8983},
  year={2024}
}

@inproceedings{zhang2024remark,
  title={REMARK-LLM: A robust and efficient watermarking framework for generative large language models},
  author={Zhang, Ruisi and Hussain, Shehzeen Samarah and Neekhara, Paarth and Koushanfar, Farinaz},
  booktitle={Proceedings of USENIX Security},
  pages={1813--1830},
  year={2024}
}

@inproceedings{ramesh2021zero,
  title={Zero-shot text-to-image generation},
  author={Ramesh, Aditya and Pavlov, Mikhail and Goh, Gabriel and Gray, Scott and Voss, Chelsea and Radford, Alec and Chen, Mark and Sutskever, Ilya},
  booktitle={Proceedings of ICML},
  pages={8821--8831},
  year={2021},
  organization={Pmlr}
}

@inproceedings{2023addconditional,
  title={Adding conditional control to text-to-image diffusion models},
  author={Zhang, Lvmin and Rao, Anyi and Agrawala, Maneesh},
  booktitle={Proceedings of CVPR},
  pages={3836--3847},
  year={2023}
}

@article{richardeau202420,
  title={The 20 questions game to distinguish large language models},
  author={Richardeau, Gurvan and Merrer, Erwan Le and Penzo, Camilla and Tredan, Gilles},
  journal={arXiv preprint arXiv:2409.10338},
  year={2024}
}

@article{pasquini2024llmmap,
  title={LLMmap: Fingerprinting for large language models},
  author={Pasquini, Dario and Kornaropoulos, Evgenios M and Ateniese, Giuseppe},
  journal={arXiv preprint arXiv:2407.15847},
  year={2024}
}

@misc{stable_diffusion_v1_4,
  title = {Stable Diffusion v1.4},
  url = {https://huggingface.co/CompVis/stable-diffusion-v1-4},
  note = {\url{https://huggingface.co/CompVis/stable-diffusion-v1-4}}
}

@misc{stable_diffusion_v2_1,
  title = {Stable Diffusion v2.1},
  url = {https://huggingface.co/stabilityai/stable-diffusion-2-1},
  note = {\url{https://huggingface.co/stabilityai/stable-diffusion-2-1}}
}

@misc{sdxl,
  title = {SDXL},
  url = {https://huggingface.co/docs/diffusers/en/using-diffusers/sdxl},
  note = {\url{https://huggingface.co/docs/diffusers/en/using-diffusers/sdxl}}
}

@misc{flux_1,
  title = {FLUX.1},
  url = {https://huggingface.co/black-forest-labs/FLUX.1-dev},
  note = {\url{https://huggingface.co/black-forest-labs/FLUX.1-dev}}
}

@misc{playground_v2_5,
  title = {Playground v2.5},
  url = {https://huggingface.co/playgroundai/playground-v2.5-1024px-aesthetic},
  note = {\url{https://huggingface.co/playgroundai/playground-v2.5-1024px-aesthetic}}
}

@misc{playground_v2,
  title = {Playground v2},
  url = {https://huggingface.co/playgroundai/playground-v2-1024px-aesthetic},
  note = {\url{https://huggingface.co/playgroundai/playground-v2-1024px-aesthetic}}
}

@misc{dalle_mini,
  title = {DALL-mini},
  url = {https://huggingface.co/spaces/dalle-mini/dalle-mini},
  note = {\url{https://huggingface.co/spaces/dalle-mini/dalle-mini}}
}

@misc{stable_diffusion_v3,
  title = {Stable Diffusion v3},
  url = {https://huggingface.co/stabilityai/stable-diffusion-3-medium},
  note = {\url{https://huggingface.co/stabilityai/stable-diffusion-3-medium}}
}

@misc{Hunyuan_DiT_v1.2,
  title = {Hunyuan DiT v1.2},
  url = {https://huggingface.co/Tencent-Hunyuan/HunyuanDiT},
  note = {\url{https://huggingface.co/Tencent-Hunyuan/HunyuanDiT}}
}

@misc{Hyper-SD,
  title = {Hyper-SD},
  url = {https://huggingface.co/ByteDance/Hyper-SD},
  note = {\url{https://huggingface.co/ByteDance/Hyper-SD}}
}

@article{Szegedy2013Intriguing,
title={Intriguing Properties of Neural Networks},
author={Christian Szegedy and Wojciech Zaremba and Ilya Sutskever and Joan Bruna and Dumitru Erhan and Ian Goodfellow and Rob Fergus},
journal={Computing Research Repository},
volume={abs/1312.6199},
year={2013}
}

@inproceedings{kou2023character,
  title={Character As Pixels: A Controllable Prompt Adversarial Attacking Framework for Black-Box Text Guided Image Generation Models.},
  author={Kou, Ziyi and Pei, Shichao and Tian, Yijun and Zhang, Xiangliang},
  booktitle={Proceedings of IJCAI},
  pages={983--990},
  year={2023}
}

@inproceedings{liu2023riatig,
  title={Riatig: Reliable and imperceptible adversarial text-to-image generation with natural prompts},
  author={Liu, Han and Wu, Yuhao and Zhai, Shixuan and Yuan, Bo and Zhang, Ning},
  booktitle={Proceedings of CVPR},
  pages={20585--20594},
  year={2023}
}

@inproceedings{zhuang2023pilot,
  title={A pilot study of query-free adversarial attack against stable diffusion},
  author={Zhuang, Haomin and Zhang, Yihua and Liu, Sijia},
  booktitle={Proceedings of CVPRW},
  pages={2385--2392},
  year={2023}
}

@article{shahgir2023asymmetric,
  title={Asymmetric Bias in Text-to-Image Generation with Adversarial Attacks},
  author={Shahgir, Haz Sameen and Kong, Xianghao and Steeg, Greg Ver and Dong, Yue},
  journal={arXiv preprint arXiv:2312.14440},
  year={2023}
}

@inproceedings{Madry2017Towards,
title={Towards Deep Learning Models Resistant to Adversarial Attacks},
author={Aleksander Madry and Aleksandar Makelov and Ludwig Schmidt and Dimitris Tsipras and Adrian Vladu},
booktitle={Proceedings of ICLR},
pages={},
year={2017}
}

@inproceedings{Carlini2017Towards,
title={Towards Evaluating the Robustness of Neural Networks},
author={Nicholas Carlini and David Wagner},
booktitle={Proceedings of S{\&}P},
year={2017}
}

@article{Chen2017ZOO:,
title={ZOO: Zeroth Order Optimization based Black-box Attacks to Deep Neural Networks without Training Substitute Models},
author={Pin-Yu Chen and Huan Zhang and Yash Sharma and Jinfeng Yi and Cho-Jui Hsieh},
journal={arXiv: Machine Learning},
volume={abs/1708.03999},
number={},
pages={15-26},
year={2017}
}

@inproceedings{gubri2024trap,
    title = "{TRAP}: Targeted Random Adversarial Prompt Honeypot for Black-Box Identification",
    author = "Gubri, Martin  and
      Ulmer, Dennis  and
      Lee, Hwaran  and
      Yun, Sangdoo  and
      Oh, Seong Joon",
    editor = "Ku, Lun-Wei  and
      Martins, Andre  and
      Srikumar, Vivek",
    booktitle = "Proceedings of ACL ",
    month = aug,
    year = "2024",
    pages = "11496--11517",
}

@inproceedings{Improving_transferability,
  title={Improving transferability of adversarial examples with input diversity},
  author={Xie, Cihang and Zhang, Zhishuai and Zhou, Yuyin and Bai, Song and Wang, Jianyu and Ren, Zhou and Yuille, Alan L},
  booktitle={Proceedings of CVPR},
  pages={2730--2739},
  year={2019}
}

@article{wu2018understanding_transferability,
  title={Understanding and enhancing the transferability of adversarial examples},
  author={Wu, Lei and Zhu, Zhanxing and Tai, Cheng and others},
  journal={arXiv preprint arXiv:1802.09707},
  year={2018}
}

@inproceedings{Liu2017Delving_Transferable,
title={Delving into Transferable Adversarial Examples and Black-box Attacks.},
author={Yanpei Liu and Xinyun Chen and Chang Liu and Dawn Song},
booktitle={Proceedings of ICLR},
year={2017}
}

@inproceedings{dong2019evading_transferable,
  title={Evading defenses to transferable adversarial examples by translation-invariant attacks},
  author={Dong, Yinpeng and Pang, Tianyu and Su, Hang and Zhu, Jun},
  booktitle={Proceedings of CVPR},
  pages={4312--4321},
  year={2019}
}

@article{AAPS,
  title={Adversarial Attacks on Parts of Speech: An Empirical Study in Text-to-Image Generation},
  author={Shahariar, GM and Chen, Jia and Li, Jiachen and Dong, Yue},
  journal={arXiv preprint arXiv:2409.15381},
  year={2024}
}

@inproceedings{yang2024sneakyprompt,
  title={Sneakyprompt: Jailbreaking text-to-image generative models},
  author={Yang, Yuchen and Hui, Bo and Yuan, Haolin and Gong, Neil and Cao, Yinzhi},
  booktitle={Proceedings of S{\&}P},
  pages={897--912},
  year={2024},
}

@inproceedings{ATM,
  title={Stable diffusion is unstable},
  author={Du, Chengbin and Li, Yanxi and Qiu, Zhongwei and Xu, Chang},
  booktitle={Proceedings of NIPS},
  volume={36},
  year={2024}
}

@inproceedings{wangunifiedShapley,
  title={A Unified Approach to Interpreting and Boosting Adversarial Transferability},
  author={Wang, Xin and Ren, Jie and Lin, Shuyun and Zhu, Xiangming and Wang, Yisen and Zhang, Quanshi},
  booktitle={Proceedings of ICLR},
  year={2021}
}

@incollection{greedy,
  title={The greedy algorithm},
  author={Jungnickel, Dieter},
  booktitle={Graphs, networks and algorithms},
  pages={129--153},
  year={1999},
  publisher={Springer}
}

@article{NSGA-II,
  title={A fast and elitist multiobjective genetic algorithm: NSGA-II},
  author={Deb, Kalyanmoy and Pratap, Amrit and Agarwal, Sameer and Meyarivan, TAMT},
  journal={IEEE transactions on evolutionary computation},
  volume={6},
  number={2},
  pages={182--197},
  year={2002},
  publisher={Ieee}
}

@article{pso,
  title={Particle swarm optimization (PSO). A tutorial},
  author={Marini, Federico and Walczak, Beata},
  journal={Chemometrics and intelligent laboratory systems},
  volume={149},
  pages={153--165},
  year={2015},
  publisher={Elsevier}
}

@inproceedings{ga,
  title={Genetic algorithm-A literature review},
  author={Lambora, Annu and Gupta, Kunal and Chopra, Kriti},
  booktitle={2019 international conference on machine learning, big data, cloud and parallel computing (COMITCon)},
  pages={380--384},
  year={2019},
  organization={IEEE}
}

@article{wang2022diffusiondb,
  title={Diffusiondb: A large-scale prompt gallery dataset for text-to-image generative models},
  author={Wang, Zijie J and Montoya, Evan and Munechika, David and Yang, Haoyang and Hoover, Benjamin and Chau, Duen Horng},
  journal={arXiv preprint arXiv:2210.14896},
  year={2022}
}

@misc{dall-e-3,
  title = {Dall-e-3},
  url = {https://openai.com/index/dall-e-3/},
  note = {\url{https://openai.com/index/dall-e-3/}}
}

@article{roubens1999axiomatic,
  title={An axiomatic approach to the concept of interaction among players in cooperative games},
  author={Roubens, Marc and Grabisch, Michel},
  journal={International Journal of Game Theory},
  volume={28},
  number={4},
  pages={547--565},
  year={1999},
  publisher={Springer}
}

@article{liu2022character,
  title={Character-level white-box adversarial attacks against transformers via attachable subwords substitution},
  author={Liu, Aiwei and Yu, Honghai and Hu, Xuming and Li, Shu'ang and Lin, Li and Ma, Fukun and Yang, Yawen and Wen, Lijie},
  journal={arXiv preprint arXiv:2210.17004},
  year={2022}
}

@inproceedings{yang2020patchattack,
  title={Patchattack: A black-box texture-based attack with reinforcement learning},
  author={Yang, Chenglin and Kortylewski, Adam and Xie, Cihang and Cao, Yinzhi and Yuille, Alan},
  booktitle={Proceedings of ECCV},
  pages={681--698},
  year={2020},
  organization={Springer}
}

@article{williams1992simple,
  title={Simple statistical gradient-following algorithms for connectionist reinforcement learning},
  author={Williams, Ronald J},
  journal={Machine learning},
  volume={8},
  number={3},
  pages={229--256},
  year={1992},
  publisher={Springer}
}

@book{kuhn1953contributions,
  title={Contributions to the Theory of Games},
  author={Kuhn, Harold W and Tucker, Albert William},
  number={28},
  year={1953},
  publisher={Princeton University Press}
}

@article{radford2019language,
  title={Language Models are Unsupervised Multitask Learners},
  author={Radford, Alec and Wu, Jeff and Child, Rewon and Luan, David and Amodei, Dario and Sutskever, Ilya},
  year={2019}
}

@article{zhang2026concept,
  title={Concept unlearning by modeling key steps of diffusion process},
  author={Zhang, Chaoshuo and Lin, Chenhao and Zhao, Zhengyu and Yang, Le and Zhang, Chong and Wang, Qian and Shen, Chao},
  journal={IEEE Transactions on Information Forensics and Security},
  year={2026},
  publisher={IEEE}
}

@article{ren2026success,
  title={On Success and Simplicity: A Second Look at Transferable Vision-Language Attack Pipeline},
  author={Ren, Y and Zhao, Z and Lin, C and Yang, B and Shen, C},
  journal={IEEE Transactions on Information Forensics and Security},
  year={2026},
  month={jul},
  day={17}
}

\begin{IEEEbiography}[{\includegraphics[width=1in]{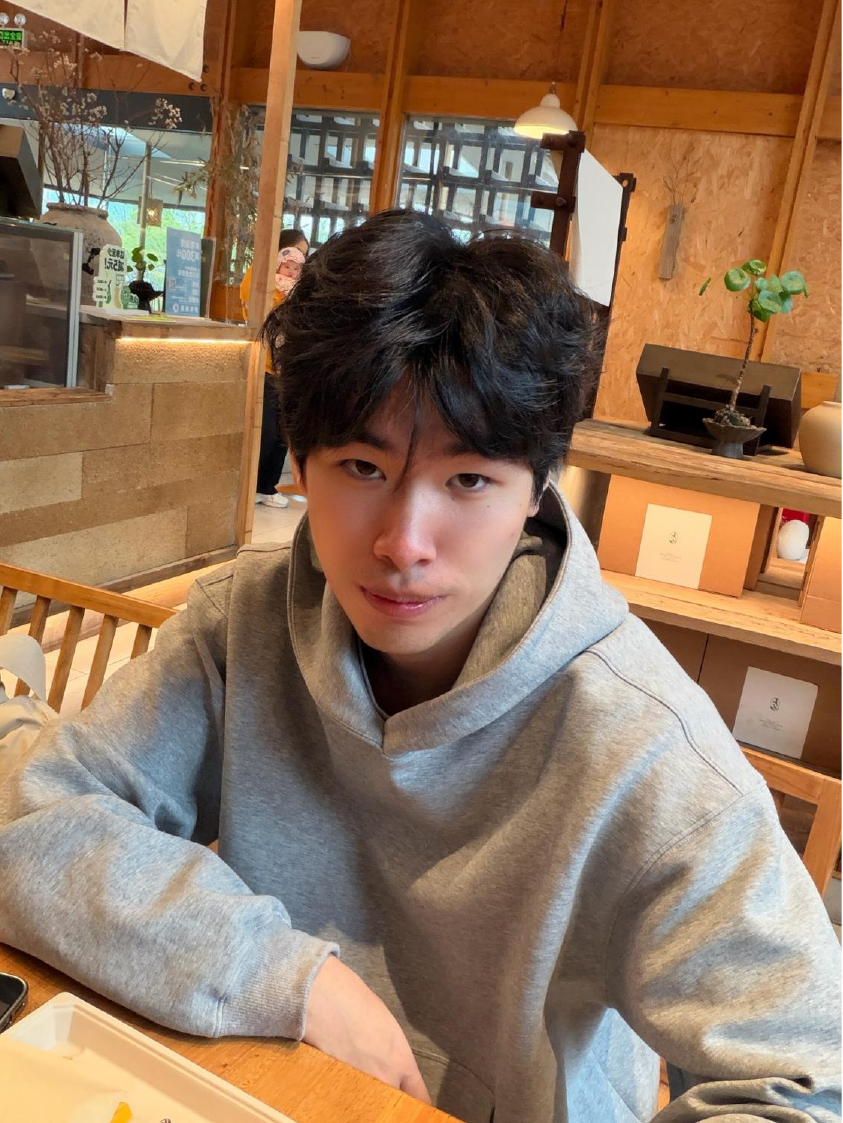}}]{Ji Guo}
 is currently a Ph.D. student at the Laboratory of Intelligent Collaborative Computing, University of Electronic Science and Technology of China (UESTC). He has authored over ten papers published in leading journals and conferences, including ICML, IEEE TDSC, and CVPR. He has also served as a reviewer for several top-tier venues, such as NeurIPS, AAAI, and ICML. His research interests include AI security, backdoor attacks, and adversarial attacks.
\end{IEEEbiography}

\begin{IEEEbiography}[{\includegraphics[width=1in]{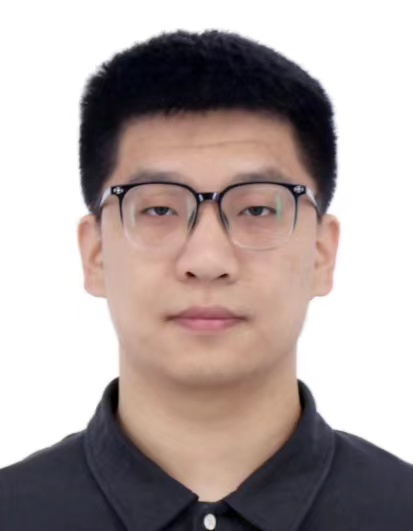}}]{Rui Zhang}
(S'20) received the B.S. degree in information security in 2020, from the University of Electronic Science and Technology of China (UESTC), where he is currently working toward the Ph.D. degree in cyber security at UESTC. His research interests include AI security and privacy-preserving deep learning.
\end{IEEEbiography}

\begin{IEEEbiography}[{\includegraphics[width=1in]{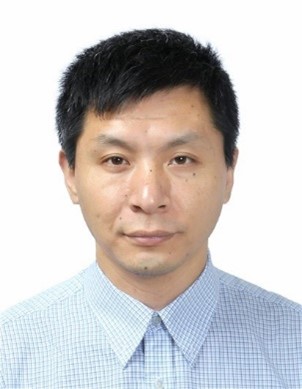}}]{Guomin~Lu}
received his Ph.D. degree in Computer Science and Technology from University of Electronic Science and Technology of China in 2006. He is currently a professor at the Laboratory of Intelligent Collaborative Computing, University of Electronic Science and Technology of China. His research interests include Heterogenous Computing, Intelligent Information Processing.
\end{IEEEbiography}

\begin{IEEEbiography}[{\includegraphics[width=1in]{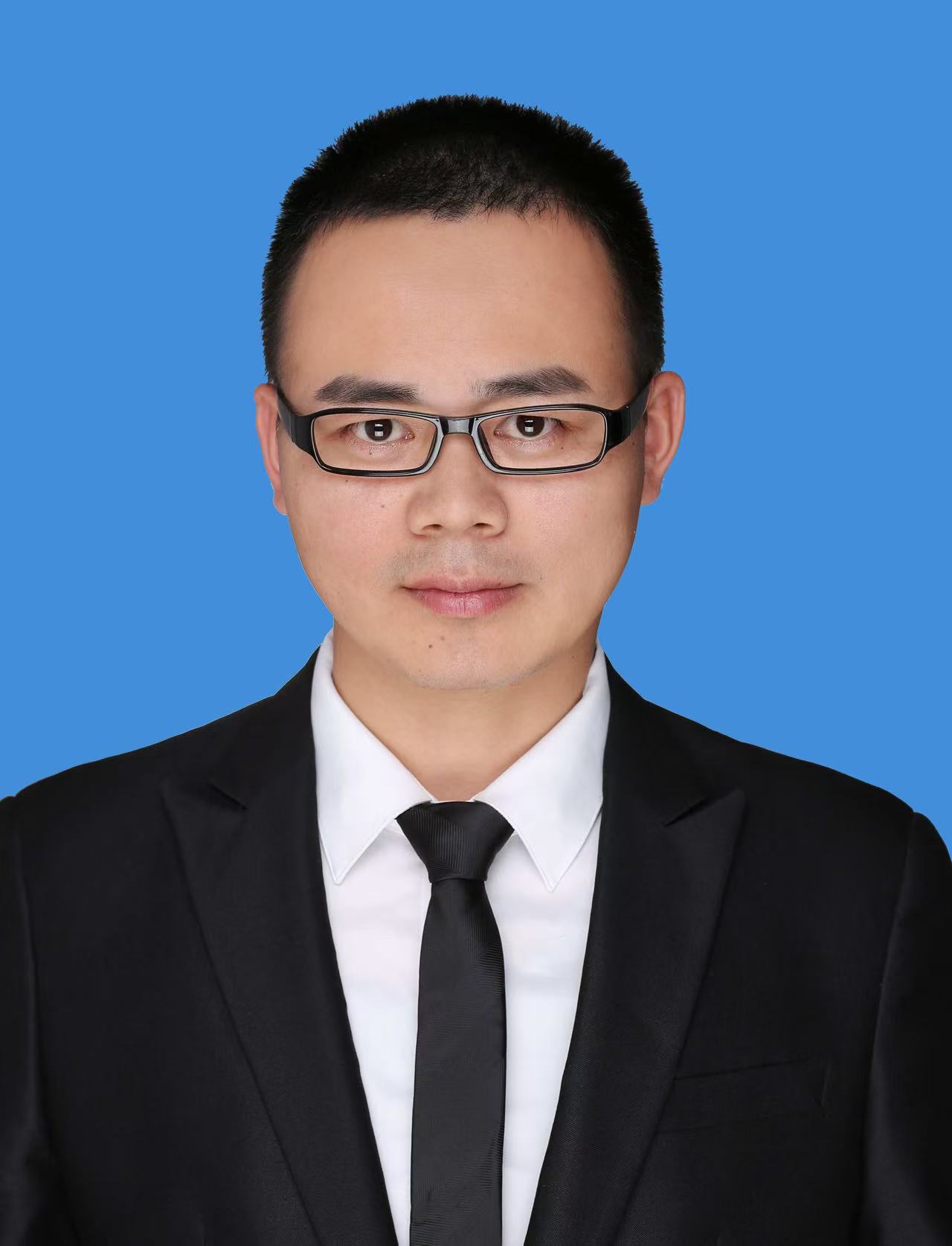}}]{Aiguo~Chen}
 received his Ph.D. degree in Signal and Information Processing from Beijing University of Posts and Telecommunications, China in 2009. He was a visiting scholar in Arizona State University, USA from Jan. 2013 to Jan. 2014. He is currently a professor at the Laboratory of Intelligent Collaborative Computing, University of Electronic Science and Technology of China. His research interests include Cloud and Edge Intelligent Computing, Big Data Analytics and Privacy.
\end{IEEEbiography}

\begin{IEEEbiography}[{\includegraphics[width=1in]{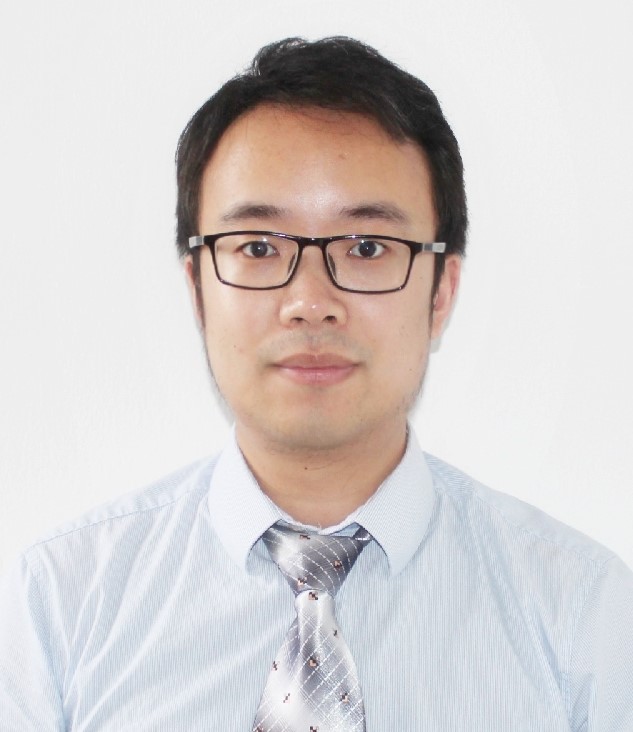}}]{Tianwei~Zhang}
received the bachelor's degree from Peking University in 2011, and the PhD degree from Princeton University in 2017. He is currently an associate professor and Provost's Chair in Computing, Nanyang Technological University. His research focuses on building efficient and trustworthy computer systems. He has been serving on the editorial boards of IEEE Transactions on Information Forensics and Security, IEEE Transactions on Multimedia, and IEEE Transactions on Circuits and Systems for VideoTechnology
\end{IEEEbiography}

\begin{IEEEbiography}[{\includegraphics[width=1in]{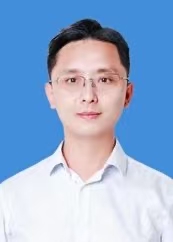}}]{Hongwei~Li}
(M’12-SM’18-F’24) is currently the Head and a Professor at Department of Infor- mation Security, School of Computer Science and Engineering, University of Electronic Sci- ence and Technology of China. He received the Ph.D. degree from University of Electronic Sci- ence and Technology of China in June 2008. He worked as a Postdoctoral Fellow at the University of Waterloo from October 2011 to October 2012. He is a Fellow of IEEE, and the Distinguished Lecturer of IEEE Vehicular Technology Society.
\end{IEEEbiography}

\begin{IEEEbiography}[{\includegraphics[width=1in]{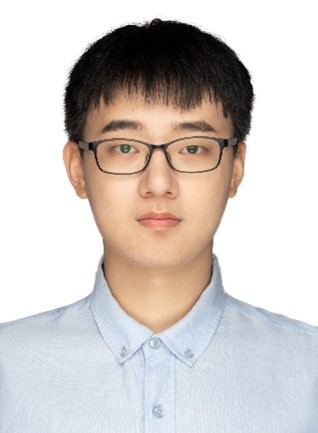}}]{Wenbo~Jiang} received his Ph.D. degree in the School of Computer Science and Engineering (School of Cyberspace Security) at the University of Electronic Science and Technology of China. He is currently an Associate Professor in the School of Computer Science and Engineering (School of Cyberspace Security) at the University of Electronic Science and Technology of China. He has published more than 60 papers in major conferences/journals, including USENIX Security, CCS, NDSS, CVPR, NeurIPS, AAAI, ICML, TDSC, TIFS, etc. He has been awarded the ACM CCS 2025 Distinguished Paper Award. He serves as area chair for many major conferences such as ICLR, NeurIPS, ICASSP, IJCNN, etc. His research interests includes AI security and data security.
\end{IEEEbiography}

\clearpage

\end{document}